\documentclass{article}

\usepackage[main,final,nonatbib]{neurips_2026}

\usepackage[utf8]{inputenc} \usepackage[T1]{fontenc}    \usepackage{hyperref}       \usepackage{url}            \usepackage{booktabs}       \usepackage{amsfonts}       \usepackage{nicefrac}       \usepackage{microtype}      \usepackage{xcolor}         
\usepackage{graphicx}
\usepackage{subcaption}
\usepackage{caption}
\usepackage{makecell}
\usepackage{pifont}
\usepackage{multirow}
\usepackage{threeparttable}
\usepackage{amsmath}
\usepackage[linesnumbered,ruled,vlined]{algorithm2e}
\usepackage{amssymb} \usepackage{array} \usepackage{wrapfig}
\usepackage[numbers,sort&compress]{natbib}

\newcommand{\gxmark}{\textcolor{gray}{\xmark}}
\newcommand{\cmark}{\ding{51}}
\newcommand{\xmark}{\ding{55}}
\newcolumntype{C}[1]{>{\centering\arraybackslash}p{#1}}
\newcolumntype{L}[1]{>{\raggedright\arraybackslash}p{#1}}

\title{ContinuLoc: Continuous Pose Inference over Neural Fields for UAV Geo-Localization}

\author{\textbf{Weikun Zhou\textsuperscript{1,2,3}} \quad
  \textbf{Jinliang Lin\textsuperscript{1}} \quad
  \textbf{Yu Zang\textsuperscript{1,2,3,}\thanks{Corresponding Author}} \quad
  \textbf{Lin Xie\textsuperscript{1,2,3}} \quad
  \textbf{Haiqing Fu\textsuperscript{1,2,3}} \\
  \textbf{Sheng Ao\textsuperscript{1,2,3}} \quad
  \textbf{Shangshu Yu\textsuperscript{4}} \quad
  \textbf{Zhiming Luo\textsuperscript{1}} \quad
  \textbf{Cheng Wang\textsuperscript{1,2,3}} \\
  {\normalfont\small \textsuperscript{1}Fujian Key Laboratory of Urban Intelligent Sensing and Computing,} \\
  {\normalfont\small Xiamen University, Xiamen, China} \\
  {\normalfont\small \textsuperscript{2}Key Laboratory of Multimedia Trusted Perception and Efficient Computing,} \\
  {\normalfont\small Ministry of Education of China, Xiamen University, Xiamen, China} \\
  {\normalfont\small \textsuperscript{3}The Belt and Road Research Institute, Xiamen University, Xiamen, China} \\
  {\normalfont\small \textsuperscript{4}School of Computer Science and Engineering, Northeastern University, Shenyang 110819, China} \\
  {\normalfont\small\texttt{zhouweikun@stu.xmu.edu.cn \quad zangyu7@xmu.edu.cn}}
}

\begin{document}

\maketitle

\vspace{-14.7pt}
\begin{abstract}
Cross-view UAV geo-localization involves three design choices: how to represent the UAV observation, how to represent the geo-referenced map, and how to infer pose from their interaction. 
The dominant paradigm organizes the map as a discretized 2DoF candidate library, coupling map representation to pose inference and thus bounding spatial localization precision by candidate sampling resolution. We decouple map representation from pose inference by encoding the geo-referenced map as a continuously queryable semantic neural field, enabling localization to be formulated as explicit optimization over a continuous 4DoF pose space --- parameterized by 2D ground position, in-plane heading, and viewing scale. Pose hypotheses are evaluated on demand using global descriptors in a shared semantic feature space, bypassing the sampling bottleneck of retrieval. We learn 4DoF pose-sensitive visual representations under full 4DoF supervision provided by ContinuScenes, a multi-city benchmark with dense near-nadir UAV imagery and complete pose annotations. Inference proceeds hierarchically, progressing from broad pose-space coverage to adaptive concentration on high-compatibility regions. Experiments demonstrate substantial gains over representative localization paradigms under progressively stricter 4DoF accuracy criteria.
Code is available at
\url{https://github.com/ZhouWeikun/ContinuLoc/}.
\end{abstract}

\section{Introduction}
Cross-view UAV geo-localization is often framed as a two-degree-of-freedom (2DoF) retrieval problem aimed at bridging severe appearance discrepancies between UAV observations and geo-referenced remote-sensing maps~\cite{Dai2022VisionBasedUS,A2025,JointRepresentationLearningAnd2022,Hu2025QueryDrivenFL,AnObliqueRobustAbsolute2024,CrowdSourcedPicturesGeo2018,VL2024,Xu2024UAVVisLocAL}. However, describing the task purely in terms of visual matching obscures its internal structure. Beyond learning discriminative visual representations, a localization system must also determine how reference-map evidence is represented and how the query pose is inferred from observation--map compatibility. This reveals three interdependent components: visual representation, map representation, and pose inference. How these components are organized jointly shapes the achievable localization accuracy, map storage requirements, and online inference cost.

Viewed through this three-component perspective, the prevailing 2DoF retrieval paradigm~\cite{Dai2022VisionBasedUS,A2025,JointRepresentationLearningAnd2022,Hu2025QueryDrivenFL,AnObliqueRobustAbsolute2024,CrowdSourcedPicturesGeo2018,VL2024,Xu2024UAVVisLocAL} has primarily emphasized learning geographically discriminative visual representations. 
The geo-referenced map is typically organized as a discrete reference library, while pose inference reduces to one-step similarity-based candidate selection, restricting pose estimates to predefined hypotheses.

For visual representation, the question is which pose cues should be preserved for localization. Under the typical near-nadir viewing setting~\cite{Xu2024UAVVisLocAL}, cross-view UAV localization can be viewed as placing a UAV observation into a geo-referenced remote-sensing map through geometric alignment.  This setting can be parameterized by a task-specific 4D localization state, consisting of 2D ground position, in-plane heading, and viewing scale. The viewing scale captures the relative ground footprint of the UAV observation with respect to the reference map, from which flight altitude can be recovered given known camera intrinsics. 
This reveals a gap between the pose information available from observation--map alignment and what the prevailing 2DoF retrieval paradigm explicitly recovers. 
Although such alignment can constrain all four pose variables, 2DoF retrieval explicitly estimates only position. 
Moreover, position-only supervision encourages observations from the same location to have similar representations regardless of heading or viewing scale, potentially suppressing the cues needed to recover these two variables. 
In this work, we instead learn 4DoF pose-sensitive visual representations, enabled by ContinuScenes, a multi-city benchmark providing full 4DoF pose annotations.
Here, we use the term 4DoF pose to describe planar similarity alignment, where viewing scale constitutes an alignment degree of freedom rather than a kinematic one~\cite{Craig1986IntroductionTR}.

While 4DoF-aware visual representations preserve richer pose cues, recovering these variables also exposes a structural limitation of the retrieval paradigm. 
\phantomsection\label{par:sampling_density_bottleneck}
Extending a discrete reference library to jointly cover position, heading, and viewing scale requires increasingly dense sampling, leading to rapidly growing map storage and indexing costs.
The underlying issue is that the reference library simultaneously serves as a representation of map evidence and an enumeration of pose hypotheses, directly coupling inference resolution to the density of stored map evidence.
From the three-component perspective, however, map representation need not be tied to the fixed set of pose hypotheses used for inference.
Indeed, a dense retrieval library is generated from the same geo-referenced map under different pose hypotheses and thus contains substantial redundancy~\cite{shannon1948mathematical}.
We exploit this redundancy by representing the geo-referenced map as a compact pose-conditioned map function~\cite{mildenhall2021nerf, kolmogorov1965three} that generates map descriptors on demand for arbitrary continuous pose hypotheses, rather than storing each sampled hypothesis independently.
For a new scene, the field is fitted solely from its geo-referenced map, without retraining the visual representation model.

\begin{figure}[htbp]
    \centering
    \includegraphics[width=0.85\textwidth]{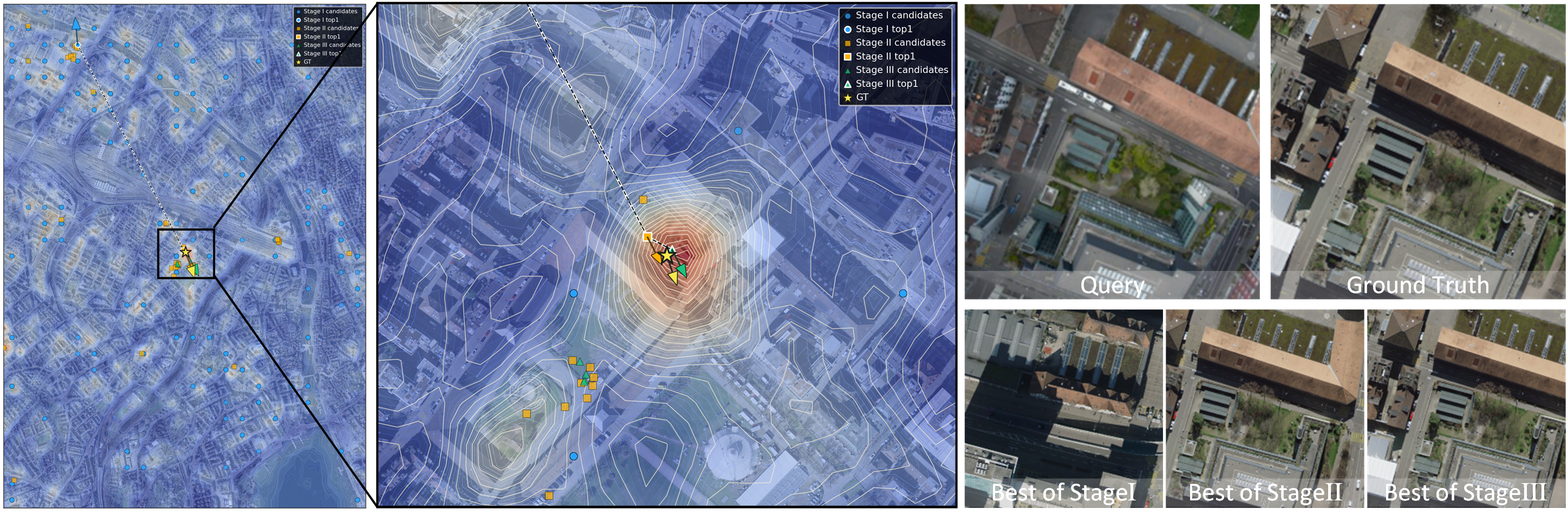}
	\caption{Illustration of ContinuLoc for UAV geo-localization.
	Left and center: visualization of the compatibility landscape, with pose candidates from the three inference stages.
	Right: the query image and reference views corresponding to
	the ground-truth pose and the highest-scoring pose at each stage.}
    \label{fig:intro_stages_zurich}
\end{figure}

This architectural decoupling, in turn, invites a rethinking of pose inference. Matching the visual representation against this pose-conditioned map function defines a continuously evaluable compatibility landscape over the pose space. Localization can thus be formulated as a global optimization problem over this landscape.
However, high-compatibility regions occupy only a small fraction of the pose space, making them costly to identify reliably through uniform sampling alone.
This gives pose inference the characteristic structure of a rare-event search problem~\cite{Rubinstein1999TheCM,Botev2013CrossEntropyOptimization}.
Motivated by distribution-adaptation principles in stochastic optimization~\cite{Ollivier2011InformationGeometricOA,Wierstra2014NaturalES}, we develop a hierarchical pose inference scheme that exploits this sparsity by progressively concentrating sampling from broad pose-space exploration toward a small number of high-compatibility regions.
Together, these designs enable pose inference to shift from one-step retrieval among fixed candidates to adaptive search over a continuous pose space, as shown in Figure~\ref{fig:intro_stages_zurich}.

In this work, we introduce ContinuLoc, a framework that rethinks UAV geo-localization from a decoupled perspective. Our contributions are threefold:
\begin{itemize}
    \item We propose a neural-field representation of geo-referenced maps that exploits redundancy in descriptors extracted from pose-conditioned reference crops, enabling compact storage and on-demand evaluation. 
    \item We formulate pose inference as a rare-event search problem and propose a hierarchical inference scheme that exploits the sparsity of high-compatibility regions to adaptively concentrate computation, avoiding dense enumeration of pose hypotheses.
    \item We construct ContinuScenes, a multi-scene UAV geo-localization
    benchmark with complete scene-level reference maps and refined 4DoF annotations.
    Experiments demonstrate the benefits of continuous pose inference and its complementarity with geometric refinement.
\end{itemize}

\section{Related Work}
\paragraph{Discrete Reference Retrieval.}
Early cross-view localization benchmarks primarily formulate the task as center-aligned retrieval~\cite{Cisneros2022ALTOAL,Dai2022VisionBasedUS} or landmark classification~\cite{Zheng2020University1652AM}. Game4Loc~\cite{Ji2024Game4LocAU} extends this paradigm to unaligned UAV localization through retrieval over discretized satellite tiles, where finite reference sampling introduces a gap between the continuous true pose and the discrete candidate set, as also evidenced by DReSS~\cite{xia2025cross}. Local geometric matching~\cite{AnObliqueRobustAbsolute2024,CrowdSourcedPicturesGeo2018,VL2024} can further refine retrieved poses beyond the reference resolution. However, such post-retrieval refinement depends on the correct region being retrieved and is typically restricted to a small top-$K$ shortlist due to computational cost~\cite{Shen2024GIMLG,Ye2025ExploringTB}. Thus, localization remains sensitive to discrete candidate coverage and initial retrieval recall.

\paragraph{Grid-Based Template Matching.}
In the broader setting of 2D-map-based localization, some methods~\cite{Sarlin2023OrienterNetVL,AirgeonetAMapGuided2024,VecmaplocnetVisionBasedUav2025,Chen2024F3LocFA} rely on dense local feature maps to evaluate pose hypotheses over discretized 3DoF spaces, producing rotation-sensitive score distributions suitable for sequential probabilistic localization~\cite{Arulampalam2002ATO,Jonschkowski2017EndtoEndLH}.
However, compared with the global-vector similarity used in descriptor retrieval, dense template matching requires aggregating similarities across multiple local features, increasing per-hypothesis evaluation costs. 
Combined with enumerative pose evaluation, this exacerbates the sampling-density bottleneck discussed in the \hyperref[par:sampling_density_bottleneck]{Introduction}.

\paragraph{Absolute Pose Regression.}
Beyond planar-map localization, 
research on 6DoF localization addressing full pose estimation in 3D environments~\cite{Zhu2024LoDLoc,Dhaouadi2025OrthoLoCU6,cheng2026pilot,Dhaouadi2026OrthoTrack}
has extensively explored absolute pose regression (APR)~\cite{NeuralRefinementForAbsolute2024,Kendall2017GeometricLF,Kendall2015PoseNetAC}, which directly predicts continuous poses from visual observations by amortizing scene-specific inference into learned network parameters. While computationally efficient, embedding the appearance-to-coordinate mapping within the regressor can complicate transfer to unseen scenes. More fundamentally, standard point-regression models do not explicitly score alternative pose hypotheses, limiting their ability to assess prediction confidence.

\paragraph{Neural Scene-Based Pose Alignment.}
Another line of 6DoF camera relocalization research builds on neural implicit map representations~\cite{Mescheder2018OccupancyNL,mildenhall2021nerf,Park2019DeepSDFLC}, which provide continuous alternatives to discrete scene storage. Existing methods typically recover poses through iterative alignment against rendered RGB or feature observations~\cite{Lin2021BARFBN,Lin2024LearningNV,Lin2020iNeRFIN}, or geometric refinement using rendered scene evidence~\cite{GS2025,FromSparseToDense2025,TheUnreasonableEffectivenessOf2024}. Although these methods provide continuous scene access, inference remains predominantly local and often requires expensive per-hypothesis rendering or alignment.
Table~\ref{tab:localization_paradigms} highlights the trade-offs among these localization paradigms in map representation and pose inference.

\begin{table}[!tbp]
    \centering
    \caption{Comparison of representative localization paradigms.}
    \label{tab:localization_paradigms}
    \scriptsize
    \setlength{\tabcolsep}{3pt}
    \renewcommand{\arraystretch}{1.2}
    \begin{tabular}{@{}L{0.23\linewidth} L{0.17\linewidth} L{0.14\linewidth} L{0.155\linewidth} L{0.22\linewidth}@{}}
        \toprule
        \textbf{Paradigm} & \textbf{Map Representation} & \textbf{Pose Hypotheses} & \textbf{Inference Strategy} & \textbf{Pose Evaluation Cost} \\
        \midrule
        Descriptor Retrieval~\cite{Dai2022VisionBasedUS,Ji2024Game4LocAU} & Global descriptors & Discrete & Enumerative & Low, vector similarity \\
        Template Matching~\cite{Sarlin2023OrienterNetVL,Chen2024F3LocFA} & Spatial local features & Discrete grid & Enumerative & Template matching \\
        APR~\cite{Kendall2015PoseNetAC,NeuralRefinementForAbsolute2024} & Regressor weights & - & Regression & Network forward pass \\
        Neural-field Alignment~\cite{Lin2020iNeRFIN,Lin2024LearningNV,GS2025} & Neural field & Continuous & Local iteration & High, render-based alignment \\
        \midrule
        \textbf{ContinuLoc} & \textbf{Neural field} & \textbf{Continuous} & \textbf{Global search} & \textbf{Low, vector similarity} \\
        \bottomrule
    \end{tabular}
\end{table}

\section{ContinuScenes: Enabling 4DoF-Aware UAV Geo-Localization}
\label{ContinuScenes_Benchmark}
\begin{figure}[!tbp] \centering \includegraphics[width=1.0\textwidth]{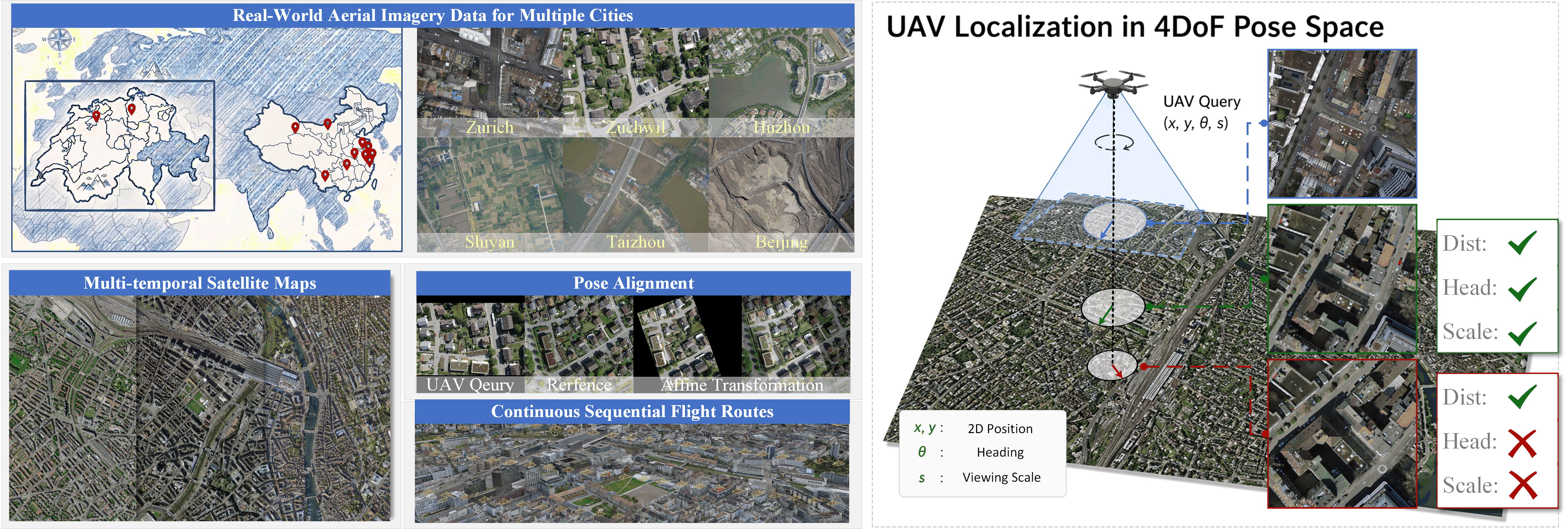} \caption{Overview of ContinuScenes and the 4DoF localization task. Left: ContinuScenes covers multiple cities with real-world UAV imagery, multi-temporal maps, and continuous flight routes. Right: The task estimates a 4DoF pose $(x, y, \theta, s)$ placing the UAV query into the reference map.} \label{fig:ContinuScenes} \end{figure}

To address the limitation of conventional 2D representation learning, we first construct ContinuScenes, as shown in Figure~\ref{fig:ContinuScenes} and Table~\ref{tab:dataset_taxonomy}, to provide the 4DoF supervision required for learning pose-sensitive visual representations.
We collect UAV images and reference maps from two sources: the publicly released UAV-VisLoc~\cite{Xu2024UAVVisLocAL} dataset for the Chinese scenes, and orthophotos from swisstopo for the Swiss scenes.
We derive refined 4DoF pose annotations from coarse sensor metadata by aligning each UAV image with its corresponding reference-map crop and extracting translation, heading, and viewing-scale corrections from the estimated 2D similarity transform.
Standardized data splits and unified evaluation protocols are also provided for this continuous 4DoF localization task.
Further details are provided in Appendix~\ref{sec:dataset_details}.

\begin{table*}[!tbp]
	\centering
	\caption{
	Comparison with representative UAV geo-localization datasets. 
}
	\label{tab:dataset_taxonomy}
	\setlength{\tabcolsep}{2pt}
	\renewcommand{\arraystretch}{1.12}
	\scriptsize
\begin{threeparttable}
	\begin{tabular}{
			@{}
			L{2.1cm}                     C{0.95cm} C{0.6cm} C{0.6cm} C{0.48cm}  C{0.9cm} C{0.8cm} C{1.2cm} C{2.15cm}     C{2.8cm}                     @{}
		}
		\specialrule{1.0pt}{0pt}{2pt}
		\multirow[c]{2}{*}[-0.9ex]{\textbf{Dataset}}
		& \multicolumn{4}{c}{\textbf{UAV Query}}
		& \multicolumn{4}{c}{\textbf{Reference Map}}
		& \multirow[c]{2}{*}[-0.9ex]{\textbf{Task Type}} \\
		\cmidrule(lr){2-5} \cmidrule(lr){6-9}
		& \textbf{Platform}
		& \textbf{Head}
		& \textbf{Scale}
		& \textbf{Seq}
		& \textbf{Regions}
		& \textbf{GNSS}
		& \textbf{Multi-date}
		& \textbf{Map-Form}
		& \\ 
		\midrule
		
		University-1652~\cite{Zheng2020University1652AM}
		& Virtual & \gxmark & \gxmark & \gxmark
		& 72 & \gxmark & \gxmark & Instance-Set
		& Landmark classification \\
		
		SUES-200~\cite{Zhu2022SUES200AM}
		& Drone & \gxmark & \gxmark & \gxmark
		& 200 & \gxmark & \gxmark & Instance-Set
		& Landmark classification \\
		
		ALTO~\cite{Cisneros2022ALTOAL}
		& Drone & \gxmark & \gxmark & \cmark
		& 1 & \cmark & \gxmark & Center-aligned Views
		& Center-aligned retrieval \\
		
		DenseUAV~\cite{Dai2022VisionBasedUS}
		& Drone & \gxmark & \cmark & \cmark
		& 14 & \cmark & \cmark & Center-aligned Views
		& Center-aligned retrieval \\
		
		GTA-UAV~\cite{Ji2024Game4LocAU}
		& Virtual & \gxmark & \cmark & \gxmark
		& 1 & \cmark & \gxmark & Tiled Map
		& Map Tile Localization \\
		
		\midrule
		Ours 
		& Drone & \cmark & \cmark & \cmark
		& 12 & \cmark & \cmark & Continuous Map
		& Continuous Pose Estimation \\
		
		\specialrule{1.0pt}{0pt}{0pt}
	\end{tabular}
\end{threeparttable}
\begin{minipage}{\linewidth}
	\scriptsize
	\emph{Note.} 
	Seq denotes whether UAV queries provide sequential observations. 
	GNSS indicates whether absolute sensor metadata, e.g., RTK, is available. 
\end{minipage}
\end{table*}

\section{ContinuLoc: Continuous Pose Inference over Neural Fields}
\subsection{Problem Formulation}

\begin{figure}[!tbp] \centering \includegraphics[width=1.0\textwidth]{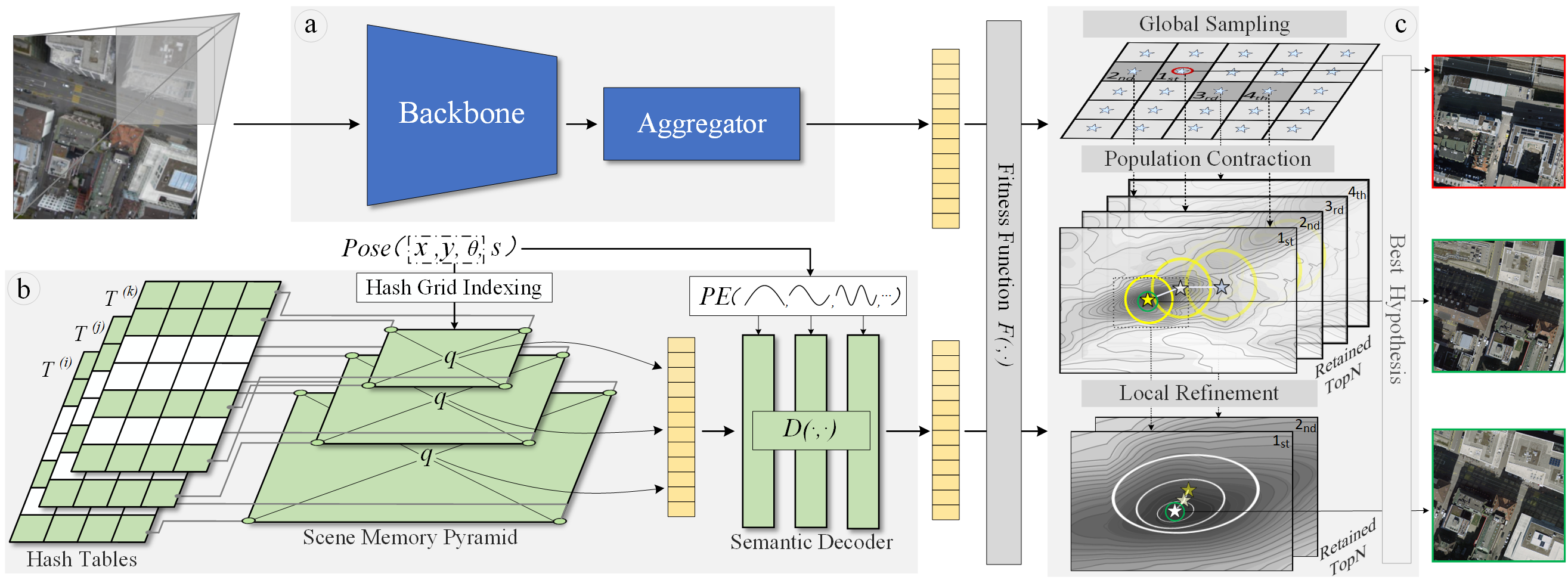} \caption{
			\textbf{Overview of ContinuLoc from the three-component perspective of UAV geo-localization.}
			(a) visual representation via a visual encoder maps the UAV query to a global descriptor;
			(b) pose-conditioned map representation via a semantic neural field provides pose-conditioned map descriptors;
			(c) pose inference searches the induced compatibility landscape from global coverage to local refinement.
			These three components interact via a shared compatibility function.}
		\label{fig:framework_overview} \label{arch} \end{figure}

Given a UAV query image $I_q$, a visual encoder parameterizes a mapping $\psi(\cdot)$ from an image to a global visual descriptor, yielding the query representation $\mathbf{z}_q=\psi(I_q)\in\mathbb{R}^d$.
On the map side, the geo-referenced map \(M\) is represented by a semantic neural field parameterized by \(\Theta_M\), which defines a mapping from the continuous pose space \(\mathcal P\) to pose-conditioned map descriptors, \(\mathbf z_M(\mathbf p)=\phi_{\Theta_M}(\mathbf p)\in\mathbb R^d\), where \(\mathbf p\in\mathcal P\).
Given these two representations, a compatibility function $\mathcal{C}(\cdot,\cdot)$ evaluates their compatibility under each pose hypothesis, yielding a scalar score $S_q(\mathbf{p})=\mathcal{C}(\mathbf{z}_q,\mathbf{z}_M(\mathbf{p}))$. Here, $\mathcal{C}(\cdot,\cdot)$ can be instantiated as cosine similarity, negative Euclidean distance, or a learnable matching function.
Evaluating this score over $\mathcal{P}$ induces a continuous compatibility landscape. Accordingly, as illustrated in Figure~\ref{fig:framework_overview}, ContinuLoc formulates UAV geo-localization as compatibility-based inference over a continuous pose space, with the objective of identifying the pose that maximizes the compatibility score:
\begin{equation}
\hat{\mathbf{p}}
=
\arg\max_{\mathbf{p}\in\mathcal{P}}
S_q(\mathbf{p}).
\label{eq:problem_statement}
\end{equation}

\subsection{Learning Pose-Sensitive Visual Representations}
Equation~\ref{eq:problem_statement} formulates localization as an argmax problem, which implicitly induces a relative ranking over pose hypotheses: the true pose should achieve a higher query--map compatibility score than competing hypotheses.
Under the near-nadir setting, each pose is parameterized as
$\mathbf p=(x,y,\theta,s)$, where $(x,y)$, $\theta$, and $s$
denote ground position, in-plane heading, and viewing scale,
respectively.
The ranking objective thus requires the visual encoder $\psi$ to produce 4DoF pose-sensitive descriptors in a shared cross-view feature space.

To learn such representations, we adopt metric learning~\cite{Hadsell2006DimensionalityRB,Schroff2015FaceNetAU}, a common paradigm in cross-view geo-localization~\cite{Dai2022VisionBasedUS,Ji2024Game4LocAU,Zhang2026InfoGeo},
using a triplet loss~\cite{Hermans2017InDefense} to encourage higher similarity between the descriptors of UAV observations and their pose-matched reference crops than between those of observations and reference crops associated with competing pose hypotheses.
Once trained, the visual encoder $\psi$ serves as a scene-generalizable component that can be reused on unseen maps without retraining. Unlike conventional approaches, we omit rotation and scale augmentations to preserve the learned visual representations' sensitivity to heading $\theta$ and viewing scale $s$.
Although we focus on 4DoF, our framework can also support traditional 2DoF localization with visual representations sensitive only to 2D position $(x,y)$.

\subsection{Neural Compression of Pose-Conditioned Semantic Fields}
Continuous pose inference requires evaluating query--map compatibility at arbitrary poses $\mathbf{p}\in\mathcal{P}$.
However, a finite reference library provides direct descriptor access only at predefined poses. Approximating arbitrary-pose access through denser sampling requires storing an increasingly large number of high-dimensional descriptors.
Drawing on neural compression
\cite{Dupont2021COINCW,Girish2023SHACIRASH},
we view the pose-indexed descriptors
$\{\mathbf{z}_M^{*}(\mathbf{p}_i)\}_{i=1}^{N}$
as discrete samples of an underlying high-dimensional, vector-valued semantic signal over the continuous pose space.
Rather than storing these sampled descriptors independently, we formulate map representation as a compression problem and learn a compact, scene-specific neural field to directly represent the underlying semantic signal:
\begin{equation}
    \phi_{\Theta_M}:\mathcal{P}\rightarrow\mathbb{R}^{d},
    \qquad
    \mathbf{p}\mapsto
    \mathbf{z}_M(\mathbf{p})=\phi_{\Theta_M}(\mathbf{p}),
\end{equation}
where $\Theta_M$ denotes the scene-specific parameters that encode the pose-conditioned descriptor signal of map $M$.
We implement this neural field using two complementary components: a multi-resolution map memory and a pose-conditioned semantic decoder.
The multi-resolution memory compactly encodes map information through shared feature parameters, while the pose-conditioned decoder combines the retrieved memory features with the full pose encoding to generate pose-conditioned map descriptors.

\paragraph{Multi-Resolution Map Memory.}
Inspired by Instant-NGP~\cite{Mller2022InstantNG}, 
we organize the map memory $\mathcal{G}$  as a multi-resolution feature pyramid, with each level implemented as an independent hash table.
At each pyramid level, the continuous coordinate space is covered by a regular grid with a level-specific resolution, and each grid vertex is mapped to an entry in a hash table by a hash encoding function.
Given a query coordinate, the entries stored at its adjacent grid vertices will be interpolated and returned as the result. 
For a pose hypothesis $\mathbf{p}=(x,y,\theta,s)$, we use its position-and-heading  component $\mathbf{q}=(x,y,\theta)$ to address the hash-grid map memory.
For level $l$, the corresponding readout can be written as:
\begin{equation}
	\mathbf{f}^{(l)}_{\mathrm{grid}}(\mathbf{q})
	=
	\operatorname{Interp}\!\left(T^{(l)},\mathbf{q}\right),
\end{equation}
where $T^{(l)}$ denotes the hash table at level $l$.
The readouts from different pyramid levels provide features with varying spatial granularities.
We concatenate them into a memory vector:
\begin{equation}
\mathbf{f}_{\mathrm{grid}}(\mathbf{q})
=
\operatorname{Concat}_{l=1}^{L}
\mathbf{f}^{(l)}_{\mathrm{grid}}(\mathbf{q}).
\end{equation}

\paragraph{Pose-Conditioned Semantic Readout.}
To perform pose-dependent interpretation of the memory vector, we introduce a 
lightweight MLP decoder that injects full-pose positional encoding (PE)~\cite{Tancik2020FourierFL} into its 
hidden transformations as a modulation signal~\cite{Niemeyer2019DifferentiableVR}.
Let 
$\tilde{\mathbf{p}}=(x,y,\cos\theta,\sin\theta,\log s)$
denote the transformed full-pose vector, where the heading is represented by 
periodic coordinates and the viewing scale is represented in log space.
We apply a sinusoidal positional encoding with $K$ frequency bands to each scalar component 
$\xi \in \tilde{\mathbf{p}}$:
\begin{equation}
	\gamma(\xi)=
	\left[
	\xi,\,
	\sin(2^0\pi \xi),\,\cos(2^0\pi \xi),\,
	\ldots,\,
	\sin(2^{K-1}\pi \xi),\,\cos(2^{K-1}\pi \xi)
	\right].
\end{equation}
The predicted map descriptor is then written as:
\begin{equation}
	\begin{aligned}
		\mathbf{e}(\mathbf{p})
		&=
		\operatorname*{Concat}_{\xi\in \tilde{\mathbf{p}}}\gamma(\xi), \\
		\mathbf{z}_M(\mathbf{p})
		&=
		\phi_{\Theta_M}(\mathbf{p})
		=
		D\!\left(
		\mathbf{f}_{\mathrm{grid}}(\mathbf{q}),
		\mathbf{e}(\mathbf{p})
		\right).
	\end{aligned}
\end{equation}

\paragraph{Semantic Field Distillation.}
Under this parameterization, learning the underlying semantic signal amounts to fitting the neural field to the sampled descriptors at their corresponding poses.
For each sampled pose hypothesis \(\mathbf{p}_i\), we extract the corresponding reference crop from the map and obtain its target map descriptor \(\mathbf{z}_M^{*}(\mathbf{p}_i)\) using the frozen visual encoder.
The field parameters are optimized to minimize the following regression objective:
\begin{equation}
\mathcal{L}_{\mathrm{map}}
=
\frac{1}{N}
\sum_{i=1}^{N}
\left\|
\phi_{\Theta_M}(\mathbf{p}_i)
-
\mathbf{z}_M^{*}(\mathbf{p}_i)
\right\|_2^2.
\end{equation}

\subsection{Hierarchical Search over the Compatibility Landscape} 
\label{Hierarchical Pose Inference in Continuous Pose Space}
Assuming that the compatibility landscape induced by the visual encoder and map field is continuous and attains its unique global maximum at the ground-truth pose, 
the region containing the true pose can be defined as follows:
\begin{equation}
\mathcal{E}_{\gamma}(q)
=
\left\{
\mathbf{p}\in\mathcal{P}
:
S_q(\mathbf{p})\ge \gamma
\right\}.
\end{equation}
The score threshold $\gamma$ is a sufficiently high value to exclude competing but distant hypotheses and restricts the region to the vicinity of the ground-truth pose.
This region occupies only a small fraction of the entire pose space.
Therefore, successful localization can be viewed as a rare-event search problem: the challenge is to efficiently reach a small set of near-optimal poses within a large pose space.

The cross-entropy method~\cite{Rubinstein1999TheCM} provides a natural principle for this rare-event search problem: adapt the proposal distribution toward high-performing samples so that high-compatibility poses become more likely to be sampled.
Standard cross-entropy optimizers characterize the search state with an explicit parameterized distribution such as a Gaussian distribution.
However, the compatibility landscape may be extremely rugged and cannot be faithfully characterized by a single simple distribution.
A direct extension is to first decompose the global pose space into many local regions and then run a separate parametric population within each plausible region~\cite{Omeradzic2023ProgressAO}.
But it comes with substantial online computation.
In fact, the sparsity of high-compatibility regions suggests that most local regions do not warrant expensive distribution adaptation.
Accordingly, we propose a three-stage, parameter-light evolutionary inference procedure~\cite{Wierstra2014NaturalES,Ollivier2011InformationGeometricOA} that progressively filters candidate regions and concentrates computational effort on the most promising ones before local refinement.

\paragraph{Stage I: Global Exploration.}
We first conduct a global evaluation over the 4DoF pose space, guided by prior knowledge, to ensure broad coverage of the compatibility landscape.
The top-scoring coarse hypotheses are retained as anchors for subsequent local refinement.

\paragraph{Stage II: Population Contraction.}
Lightweight particle populations are initialized around the retained anchors, one for each local candidate region.
We evolve these populations through repeated selection and resampling.
Within each population, top-scoring particles are selected as elites; their aggregate location re-centers the population toward the local high-compatibility region, while radius contraction progressively reduces the search range.
Across populations, persistently weak regions are eliminated, whereas strong populations are retained in parallel to preserve competing hypotheses.
This stage implements an evolutionary contraction process that preserves multi-modal uncertainty while reducing the initial coarse regions to a compact set of retained modes.

\paragraph{Stage III: Local Refinement.}
After population contraction, the search has been narrowed to a small number of plausible local modes, shifting the inference problem from multi-modal exploration to local exploitation. 
Since only a few modes remain, it becomes affordable to use covariance-adaptive evolutionary refinement~\cite{Hansen2016TheCE}, adaptively exploring the local compatibility landscape around each retained mode.
The final prediction is thus selected as the highest-scoring pose across all covariance-adaptive refinements.
Further algorithmic details are provided in Appendix~\ref{sec:hierarchical_search_details}.

\section{Experiments}
\label{sec:Experiments}
\paragraph{Datasets.}
We evaluate on four representative ContinuScenes scenes (Section~\ref{ContinuScenes_Benchmark}): Taizhou03, Taizhou04, Zurich, and Zuchwil.
Taizhou03/04 cover larger areas with higher-altitude UAV observations, while Zurich 
and Zuchwil are smaller urban scenes with lower-altitude observations, enabling 
evaluation of both global hypothesis search and fine-grained local pose estimation.
Viewing scale is supervised by the scene-normalized label 
$s^{*}=h^{*}/\bar{h}$, where $h^{*}$ is the query footprint and $\bar{h}$ is the 
average UAV footprint of the scene.
Unless otherwise stated, the main paper reports same-area results.
Detailed dataset statistics and cross-area results are provided in Table~\ref{tab:dataset_summary} and Table~\ref{tab:gim_refinement_progressive_recall_segment} in the appendix.

\paragraph{Evaluation Metric.}
We evaluate localization accuracy by measuring distance, heading, and scale errors.
For a predicted pose $\hat{\mathbf{p}}=(\hat{x},\hat{y},\hat{\theta},\hat{s})$ and ground-truth $\mathbf{p}^{*}=(x^{*},y^{*},\theta^{*},s^{*})$:
\begin{align}
	e_d &= \left\|(\hat{x},\hat{y})-(x^{*},y^{*})\right\|_2, \\
	e_h &= \min\left(|\hat{\theta}-\theta^{*}|,\ 360^{\circ}-|\hat{\theta}-\theta^{*}|\right), \\
	e_s &= \max\left(\frac{\hat{s}}{s^{*}},\frac{s^{*}}{\hat{s}}\right).
\end{align}
We report nested recall R@1 under three cumulative criteria: distance only ($D$), distance and heading ($D\&H$), and all three ($D\&H\&S$).
Recall values are reported as percentages; distance and heading errors are measured in meters and degrees, respectively, and scale errors are dimensionless ratios.
We use two median-error metrics in this paper: overall median errors computed over 
all queries, and Median@Success computed only over queries satisfying the corresponding 
recall criterion.

\subsection{Implementation Details}
We use DINOv2~\cite{Oquab2023DINOv2LR} as the visual backbone and fine-tune it with parallel adapters following SeLaVPR~\cite{Lu2024TowardsSA}.
SALAD~\cite{Izquierdo2023OptimalTA} is adopted as the global aggregator to produce 1024-dimensional descriptors for both UAV observations and pose-conditioned reference crops.
For each scene, the semantic neural field is trained to output 1024-dimensional descriptors in the same embedding space.
Detailed architecture and training hyperparameters are provided in Appendix~\ref{sec:implementation}.

\subsection{Localization Performance across Scenes}
\begin{table*}[tbp]
	\centering
	\caption{Comparison of representative localization paradigms on four scenes.}
	\label{tab:main_res_interval}
	\setlength{\tabcolsep}{3.0pt}
	\renewcommand{\arraystretch}{0.98}
	\scriptsize
	\begin{tabular}{llcccccccccccc}
		\toprule
		& & \multicolumn{6}{c}{Taizhou03}
		& \multicolumn{6}{c}{Taizhou04} \\
		\cmidrule(lr){3-8} \cmidrule(lr){9-14}
		& & \multicolumn{3}{c}{R@1 (100\,m, 10$^\circ$, $1.2\times$) $\uparrow$}
		& \multicolumn{3}{c}{Median@Success $\downarrow$}
		& \multicolumn{3}{c}{R@1 (100\,m, 10$^\circ$, $1.2\times$) $\uparrow$}
		& \multicolumn{3}{c}{Median@Success $\downarrow$} \\
		\cmidrule(lr){3-5} \cmidrule(lr){6-8}
		\cmidrule(lr){9-11} \cmidrule(lr){12-14}
		Paradigm & Method
		& D & D\&H & D\&H\&S & Dist & Head & Scale
		& D & D\&H & D\&H\&S & Dist & Head & Scale \\
		\midrule
		
		\multirow{2}{*}{2DoF Ret.}
		& DenseUAV~\cite{Dai2022VisionBasedUS}
		& 30.87 & -- & -- & 71.99 & -- & -- 
		& 20.42 & -- & -- & 78.13 & -- & -- \\
		& QDFL~\cite{Hu2025QueryDrivenFL}
		& 32.21 & -- & -- & 72.53 & -- & -- 
		& 18.31 & -- & -- & 76.87 & -- & -- \\
		
		\midrule
		
		\multirow{3}{*}{4DoF Ret.}
		& NetVLAD~\cite{Arandjelovi2015NetVLADCA}
		& 30.20 & 16.78 & 15.44 & 75.32 & 3.63 & 1.060
		& 17.61 & 11.97 & 11.97 & 76.76 & 5.40 & 1.044 \\
		& SALAD~\cite{Izquierdo2023OptimalTA}
		& 28.86 & 19.46 & 18.79 & 75.32 & 4.82 & 1.086
		& 14.08 & 9.86 & 9.86 & 74.59 & \textbf{3.83} & 1.044 \\
		& SeLaVPR~\cite{Lu2024TowardsSA}
		& 28.86 & 21.48 & 20.13 & 72.67 & \textbf{3.09} & 1.083
		& 17.61 & 11.97 & 11.97 & 78.13 & 5.40 & 1.059 \\
		
		\midrule
		
		\multirow{1}{*}{APR}
		& PoseNet~\cite{Kendall2015PoseNetAC}
		& 1.34 & 0.00 & 0.00 & -- & -- & -- 
		& 1.41 & 0.00 & 0.00 & -- & -- & -- \\
		
		\midrule
		
		\multirow{1}{*}{\makecell[l]{Cont. inf.}}
		& Ours
		& \textbf{91.28} & \textbf{81.21} & \textbf{81.21} & \textbf{30.98} & 4.27 & \textbf{1.047}
		& \textbf{83.10} & \textbf{76.76} & \textbf{76.76} & \textbf{48.53} & 4.03 & \textbf{1.020} \\
		
		\specialrule{0.08em}{0.4em}{0.4em}
		
		& & \multicolumn{6}{c}{Zurich}
		& \multicolumn{6}{c}{Zuchwil} \\
		\cmidrule(lr){3-8} \cmidrule(lr){9-14}
		& & \multicolumn{3}{c}{R@1 (50\,m, 10$^\circ$, $1.2\times$) $\uparrow$}
		& \multicolumn{3}{c}{Median@Success $\downarrow$}
		& \multicolumn{3}{c}{R@1 (50\,m, 10$^\circ$, $1.2\times$) $\uparrow$}
		& \multicolumn{3}{c}{Median@Success $\downarrow$} \\
		\cmidrule(lr){3-5} \cmidrule(lr){6-8}
		\cmidrule(lr){9-11} \cmidrule(lr){12-14}
		Paradigm & Method
		& D & D\&H & D\&H\&S & Dist & Head & Scale
		& D & D\&H & D\&H\&S & Dist & Head & Scale \\
		\midrule
		
		\multirow{2}{*}{2DoF Ret.}
		& DenseUAV~\cite{Dai2022VisionBasedUS}
		& 19.48 & -- & -- & 24.04 & -- & -- 
		& 23.01 & -- & -- & 24.75 & -- & -- \\
		& QDFL~\cite{Hu2025QueryDrivenFL}
		& 27.31 & -- & -- & 19.82 & -- & -- 
		& 29.76 & -- & -- & 25.83 & -- & -- \\
		
		\midrule
		
		\multirow{3}{*}{4DoF Ret.}
		& NetVLAD~\cite{Arandjelovi2015NetVLADCA}
		& 27.31 & 13.45 & 8.82 & 24.28 & 3.40 & 1.112
		& 32.18 & 16.26 & 12.11 & 27.95 & 3.95 & 1.053 \\
		& SALAD~\cite{Izquierdo2023OptimalTA}
		& 26.89 & 11.34 & 6.30 & 19.66 & 3.94 & \textbf{1.055}
		& 25.95 & 13.15 & 10.38 & 20.59 & 4.48 & 1.064 \\
		& SeLaVPR~\cite{Lu2024TowardsSA}
		& 31.09 & 19.33 & 11.34 & 25.36 & 3.94 & 1.085
		& 30.10 & 23.18 & 19.03 & 25.78 & 3.71 & 1.084 \\
		
		\midrule
		
		\multirow{1}{*}{APR}
		& PoseNet~\cite{Kendall2015PoseNetAC}
		& 7.56 & 2.94 & 2.94 & 32.51 & 3.10 & 1.063
		& 10.03 & 3.81 & 3.81 & 23.23 & 4.40 & \textbf{1.013} \\
		
		\midrule
		
		\multirow{1}{*}{\makecell[l]{Cont. inf.}}
		& Ours
		& \textbf{89.92} & \textbf{81.93} & \textbf{70.59} & \textbf{9.31} & \textbf{2.86} & 1.079
		& \textbf{76.12} & \textbf{61.94} & \textbf{61.25} & \textbf{10.79} & \textbf{2.72} & 1.070 \\
		
		\bottomrule
	\end{tabular}
\end{table*}

\begin{table*}[tbp]
	\centering
	\caption{Effect of GIM$_{dkm}$-based local refinement on progressive recall.}
	\label{tab:gim_refinement_progressive_recall}
	\setlength{\tabcolsep}{2.6pt}
	\renewcommand{\arraystretch}{0.98}
	\scriptsize
	\begin{tabular}{lcccccccccccc}
		\toprule
		& \multicolumn{3}{c}{Taizhou03}
		& \multicolumn{3}{c}{Taizhou04}
		& \multicolumn{3}{c}{Zurich}
		& \multicolumn{3}{c}{Zuchwil} \\
		\cmidrule(lr){2-4}
		\cmidrule(lr){5-7}
		\cmidrule(lr){8-10}
		\cmidrule(lr){11-13}
		Method
		& D & D\&H & D\&H\&S
		& D & D\&H & D\&H\&S
		& D & D\&H & D\&H\&S
		& D & D\&H & D\&H\&S \\
		\midrule
		
		QDFL~\cite{Hu2025QueryDrivenFL}
		& 86.58 & 86.58 & 86.58
		& 73.24 & 73.24 & 72.54
		& 48.74 & 48.74 & 39.50
		& 50.87 & 49.48 & 48.79 \\
		
		SeLaVPR~\cite{Lu2024TowardsSA}
		& 67.11 & 67.11 & 67.11
		& 64.79 & 64.79 & 64.08
		& 48.74 & 48.32 & 43.28
		& 47.75 & 46.37 & 46.02 \\
		
		Ours
		& \textbf{99.33} & \textbf{99.33} & \textbf{99.33}
		& \textbf{88.03} & \textbf{88.03} & \textbf{87.32}
		& \textbf{91.60} & \textbf{91.18} & \textbf{81.93}
		& \textbf{81.31} & \textbf{78.55} & \textbf{77.16} \\
		
		\bottomrule
	\end{tabular}
\end{table*}

Table~\ref{tab:main_res_interval} compares representative localization paradigms, including 2DoF retrieval, discrete 4DoF retrieval, APR, and our continuous inference framework, under increasingly strict pose constraints.
Although APR offers a compact map representation, its localization accuracy is substantially lower than that of explicit retrieval. In contrast, our framework achieves the highest R@1 under all three criteria across all four scenes.
Table~\ref{tab:gim_refinement_progressive_recall} further isolates the role of matching-based local refinement using GIM~\cite{Shen2024GIMLG} applied to the top-5 retained candidates.
Local refinement benefits both retrieval-based methods and our framework. 
Notably, before using GIM, our framework already outperforms GIM-refined retrieval baselines on three of the four scenes under the full $D\&H\&S$ criterion.
This indicates that the benefits of local matching remain fundamentally bounded by the quality and coverage of the initial hypotheses.

\subsection{Dense Retrieval versus Continuous Inference} 
\label{sec:Comparison with Dense Retrieval}

\begin{figure}[htbp]
	\centering
	\begin{minipage}[t]{0.39 \linewidth}
		\vspace{0pt}
		\centering
		\includegraphics[width=\linewidth]{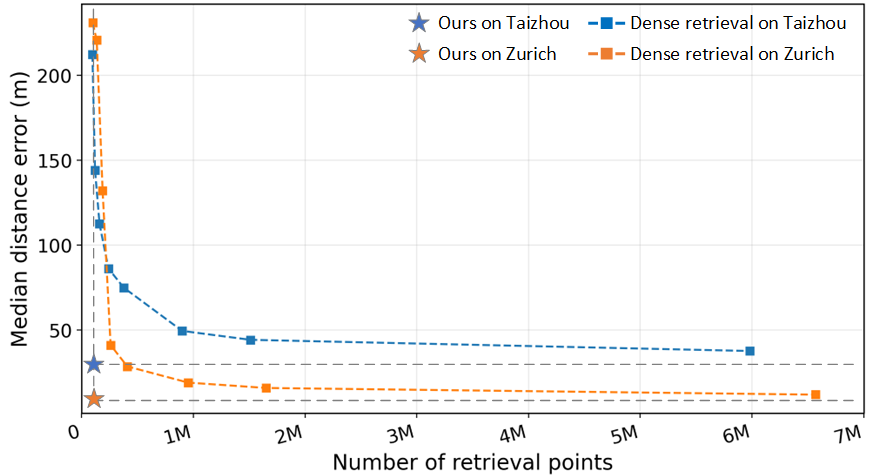}
	\end{minipage}\hfill\begin{minipage}[t]{0.55\linewidth}
	\vspace{0pt}
	\centering
	\vspace{2mm} \begingroup
	\fontsize{5.8pt}{6.4pt}\selectfont
	\setlength{\tabcolsep}{1.6pt}
	\renewcommand{\arraystretch}{0.92}
	\resizebox{\linewidth}{!}{
		\begin{tabular}{lccccccc}
			\toprule
			Metric
			& \multicolumn{5}{c}{Dense retrieval only}
			& w/ S-II
			& Full \\ \midrule
			
			InitPts
			& \textbf{98k} & 259k & 408k & 954k & 6.6M
			& \textbf{98k} & \textbf{98k} \\
			Storage (GB)$\downarrow$
			& \textbf{0.4} & 1.1 & 1.7 & 3.9 & 26.9
			& 2.8 & 2.8 \\
			Time (ms)$\downarrow$
			& \textbf{1.3} & 3.6 & 5.7 & 33.3 & 350.6
			& 98.0 & 336.4 \\
			MedErr (m)$\downarrow$
			& 231.1 & 41.1 & 28.5 & 19.0 & 11.9
			& 14.5 & \textbf{10.2} \\
			R@1 $\uparrow$
			& 8.0 & 19.3 & 24.0 & 30.3 & 36.1
			& 63.0 & \textbf{70.6} \\
			\bottomrule
		\end{tabular}
	}
	\endgroup
\end{minipage}
	\caption{
	Accuracy-efficiency trade-off between dense retrieval and our continuous inference framework.
	Left: Median localization error versus reference-library density.
	Right: Storage, runtime, and localization accuracy for dense retrieval and our inference stages.
	}
	\label{fig:dense_retrieval_tradeoff}
\end{figure}

A natural question is whether retrieval-based localization can be made sufficiently accurate simply by increasing the sampling density of the reference library.
We therefore compare dense-retrieval libraries with different numbers of reference 
points.
As shown in Figure~\ref{fig:dense_retrieval_tradeoff}, denser retrieval libraries initially reduce localization error, but they exhibit clear diminishing marginal returns as the number of reference points continues to grow.
This reflects a structural limitation of dense retrieval: improving precision requires a disproportionately larger reference library.
In contrast, our approach achieves a comparable median localization error after Stage~II using substantially less storage and runtime. The optional Stage~III refinement further improves localization accuracy at additional computational cost.
And this efficiency advantage is expected to become more pronounced in higher-dimensional pose spaces (e.g., 6DoF).

\subsection{Ablation Study}
\begin{figure}[!htbp] \centering \includegraphics[width=0.825\textwidth]{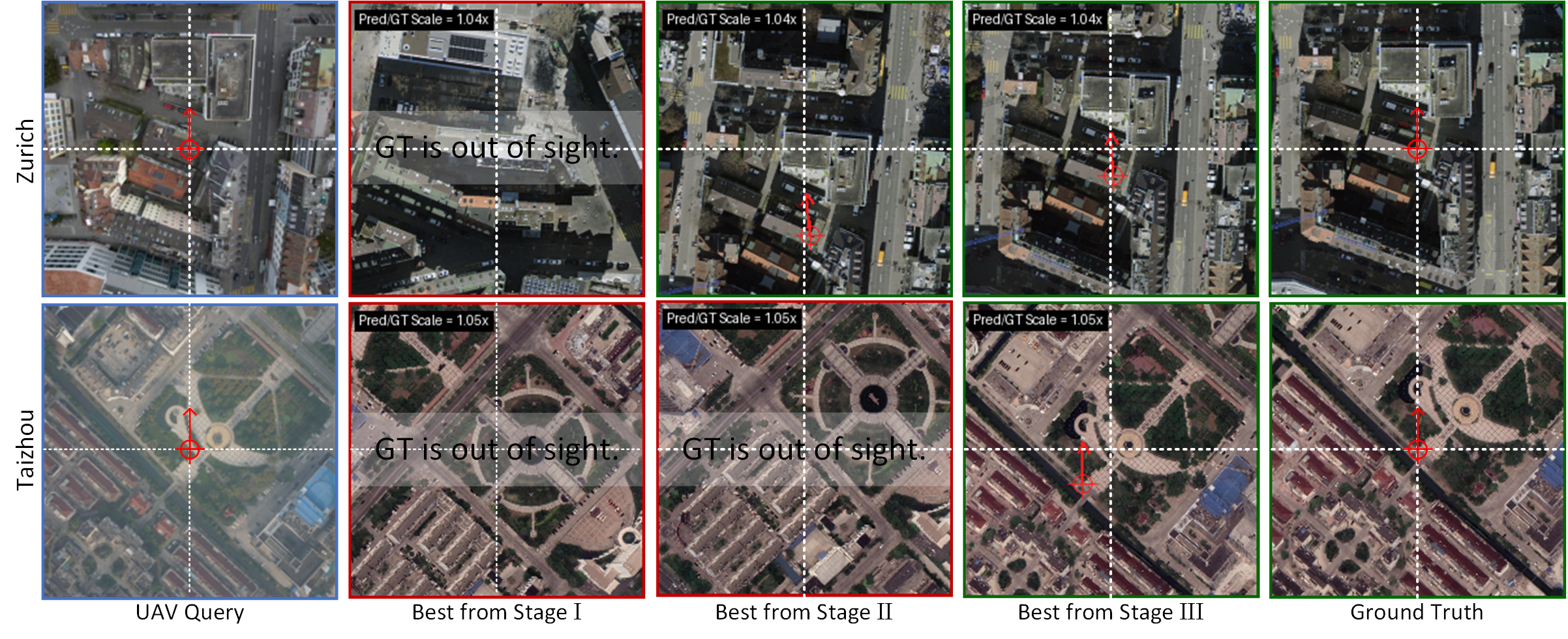} \caption{Qualitative visualization of stage-wise pose refinement.
The red arrows indicate the ground-truth query positions and orientations in the coordinate frames of the reference views corresponding to the predicted poses.
} \label{fig:stagewise_progression} \end{figure}

\begin{table*}[!htbp]
	\centering
	\caption{Stage-wise progression of the inference algorithm.}
	\label{tab:stagewise_progression}
	\setlength{\tabcolsep}{3.2pt}
	\renewcommand{\arraystretch}{0.98}
	\scriptsize
	
	\begin{tabular}{lcccccccccccc}
		\toprule
		& \multicolumn{6}{c}{Taizhou03}
		& \multicolumn{6}{c}{Zurich} \\
		\cmidrule(lr){2-7} \cmidrule(lr){8-13}
		& \multicolumn{3}{c}{R@1 (100\,m, 10$^\circ$, $1.2\times$) $\uparrow$}
		& \multicolumn{3}{c}{Median $\downarrow$}
		& \multicolumn{3}{c}{R@1 (50\,m, 10$^\circ$, $1.2\times$) $\uparrow$}
		& \multicolumn{3}{c}{Median $\downarrow$} \\
		\cmidrule(lr){2-4} \cmidrule(lr){5-7}
		\cmidrule(lr){8-10} \cmidrule(lr){11-13}
		Stage
		& D & D\&H & D\&H\&S & Dist & Head & Scale
		& D & D\&H & D\&H\&S & Dist & Head & Scale \\
		\midrule
		Global Exploration
		& 30.20 & 24.16 & 24.16 & 160.83 & 12.94 & 1.044
		& 31.93 & 22.27 & 20.17 & 115.11 & 29.97 & 1.102 \\
		Population Contraction
		& 89.26 & 75.17 & 75.17 & 43.04 & 5.13 & 1.044
		& 87.82 & 74.79 & 63.03 & 14.55 & 4.42 & 1.102 \\
		Local Refinement
		& 91.28 & 81.21 & 81.21 & 34.45 & 5.00 & 1.044
		& 89.92 & 81.93 & 70.59 & 10.15 & 3.55 & 1.102 \\
		\bottomrule
	\end{tabular}
\end{table*}

\paragraph{Stage-wise Inference Analysis.}
To evaluate how each stage of our hierarchical inference algorithm contributes to the final localization result, we take the top-scoring pose hypothesis produced at each stage and compare it with the ground-truth pose.
Stage I evaluates a sparse set of pose hypotheses, providing only a coarse sampling of the entire landscape. Consequently, the highest-scoring sampled hypothesis may not identify the most promising region across the underlying landscape. As shown in Table~\ref{tab:stagewise_progression} and Figure~\ref{fig:stagewise_progression}, Stages II and III progressively resample and refine the retained candidate regions, providing denser local evaluations that can revise their relative ranking and yield more accurate pose estimates.
The Taizhou example is deliberately selected to illustrate that visually consistent localization can still be achieved despite slight residual errors in the ground-truth annotation.

\paragraph{Pose Modulation in the Semantic Field.}
\begin{figure*}[!htbp]
	\centering
	
	\begin{minipage}[t]{0.365\textwidth}
		\vspace{0pt}
		\centering
		\includegraphics[
		width=\linewidth,
		]{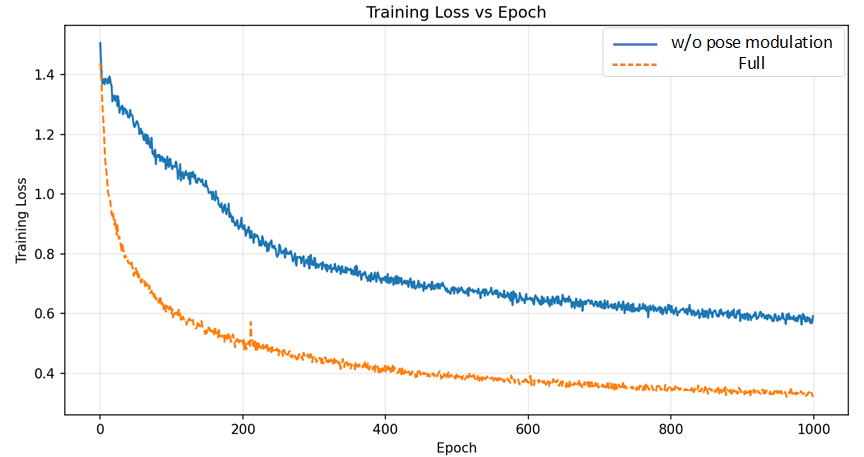}
	\end{minipage}
	\hfill
	\begin{minipage}[t]{0.56\textwidth}
		\vspace{0pt}
		\centering
		\setlength{\tabcolsep}{3.0pt}
		\renewcommand{\arraystretch}{0.96}
		\scriptsize
		\vspace{0.5mm} \begin{tabular}{@{}llcccc@{}}
			\toprule
			& Stage 
			& D $\uparrow$
			& D\&H $\uparrow$
			& D\&H\&S $\uparrow$
			& $\Delta$D\&H\&S $\uparrow$ \\
			\midrule
			w/o pose-PE mod.
			& S-I
			& 28.15 & 15.55 & 13.03 & -- \\
			& S-II
			& 33.19 & 20.17 & 17.65 & +4.62 \\
			& S-III
			& 33.19 & 22.69 & 20.17 & +7.14 \\
			\midrule
			Full
			& S-I
			& 31.93 & 22.27 & 20.17 & -- \\
			& S-II
			& 87.82 & 74.79 & 63.03 & +42.86 \\
			& S-III
			& \textbf{89.92} & \textbf{81.93} & \textbf{70.59} & \textbf{+50.42} \\
			\bottomrule
		\end{tabular}
	\end{minipage}
	\caption{
		Effect of pose modulation in the semantic field.
		Left: map-field training loss with and without pose modulation.
		Right: stage-wise R@1 under $D$, $D\&H$, and $D\&H\&S$ criteria.
	}
	\label{fig:pose_pe_ablation}
\end{figure*}

Figure~\ref{fig:pose_pe_ablation} shows the effect of pose-PE modulation.
Removing pose modulation only moderately affects coarse-stage recall, but substantially limits the gains from later-stage refinement.
The training curves suggest that this is due to weaker field learnability, as the unmodulated field converges more slowly and reaches a higher fitting loss.
Thus, pose-PE modulation helps produce pose-conditioned map descriptors that better support continuous refinement.

\subsection{Geometric Analysis of Compatibility Landscapes for Visual Localization}
\begin{figure}[!ht]
\centering
\includegraphics[width=0.95\textwidth]{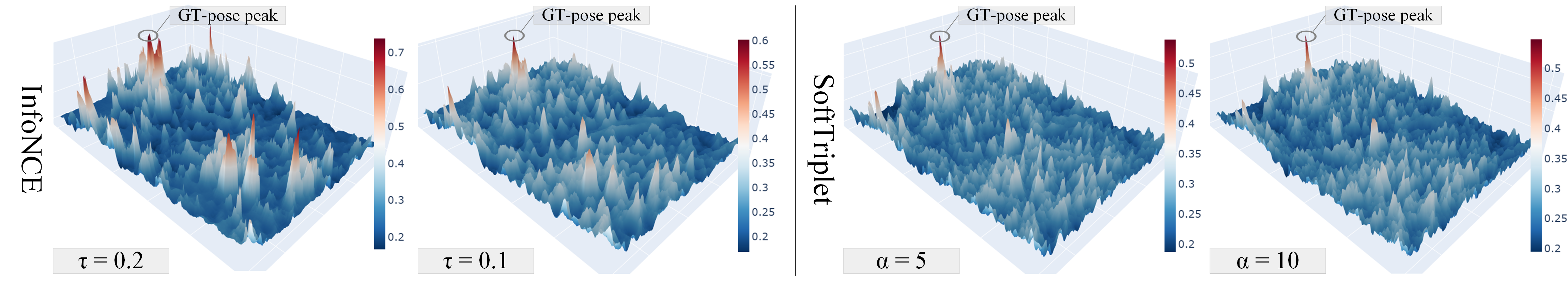}
\caption{Compatibility landscapes induced by different loss configurations. Circles mark the ground-truth poses.}
\label{fig:loss_landscape_1row}
\end{figure}

\begin{wraptable}{r}{0.38\textwidth}
\vspace{-\intextsep}
\centering
\captionsetup{
    font=footnotesize,
    skip=4pt,
}
\caption{R@1 at different sampling densities with different loss parameters.}
\label{tab:loss_parameter_sampling}
\footnotesize
\setlength{\tabcolsep}{3.5pt}
\renewcommand{\arraystretch}{1.08}
\begin{tabular}{@{}lcccc@{}}
\toprule
\multirow{2}{*}{Parameter}
& \multicolumn{4}{c}{R@1 at Sampling Density $\uparrow$} \\
\cmidrule(lr){2-5}
& $1\times$ & $1.31\times$ & $2.56\times$ & $64\times$ \\
\midrule
\multicolumn{5}{@{}l}{InfoNCE} \\[-2pt]
$\tau=0.2$ & 21.76 & 29.71 & 47.70 & 66.53 \\
$\tau=0.1$ & \textbf{46.03} & \textbf{51.88} & 71.97 & 98.33 \\
\midrule
\multicolumn{5}{@{}l}{SoftTriplet} \\[-2pt]
$\alpha=5$ & 38.08 & 40.59 & 67.78 & \textbf{100.00} \\
$\alpha=10$ & 39.33 & 43.93 & \textbf{74.06} & \textbf{100.00} \\
\bottomrule
\end{tabular}
\par\vspace{\dimexpr-\intextsep+3pt\relax}
\end{wraptable}

Viewing localization as continuous optimization over a
continuous landscape provides a geometric perspective for analyzing the localization process.
With heading and viewing scale fixed to their ground-truth values, we examine the resulting 2D compatibility landscapes and R@1 under varying spatial sampling densities.
Table~\ref{tab:loss_parameter_sampling} shows that InfoNCE with $\tau=0.2$ achieves only 66.53\% R@1 even at the densest setting.
This failure is geometrically reflected in Figure~\ref{fig:loss_landscape_1row}, where its corresponding landscape exhibits multiple prominent competing responses.
In contrast, the other three configurations approach saturated R@1 as candidate density increases, consistent with their more clearly defined ground-truth-associated peaks in Figure~\ref{fig:loss_landscape_1row}.
This suggests that finite candidate sampling provides only a sparse observation of the continuous landscape, and R@1 under a given sampling density may therefore not faithfully reflect its localization potential.

These observations suggest a broader perspective on visual localization. Conventional approaches have primarily focused on improving the discriminability of visual representations, which implicitly shapes the resulting compatibility landscape. From an optimization perspective, however, a desirable landscape should not only exhibit a well-defined structure that distinguishes the correct pose, but also be sufficiently observable under finite sampling.

\section{Conclusion}
This paper revisits UAV geo-localization through a three-component perspective encompassing visual representation, map representation, and pose inference. Building on this perspective, ContinuLoc learns 4DoF pose-sensitive observation representations, encodes geo-referenced maps as compact, continuously queryable pose-conditioned neural fields, and performs hierarchical search over the resulting observation--map compatibility landscape. 
By decoupling map representation from pose inference, ContinuLoc shifts UAV geo-localization from discrete retrieval to continuous-space optimization.
Experiments demonstrate consistent improvements over discrete retrieval and absolute pose regression, while providing stronger initial hypotheses for subsequent local matching refinement.
More broadly, our analysis suggests that localization depends not only on feature discriminability, but also on whether the induced compatibility landscape exhibits a clear geometric structure that makes the correct pose readily identifiable.

\section{Acknowledgements}
This work was supported by the National Natural Science Foundation of China (Grant Nos.~62471415 and 62401225), the Natural Science Foundation of Fujian Province (Grant Nos.~2023J01004, 2025J0141, and 2024J01115), and the Xiamen Municipal Natural Science Foundation (Grant No.~3502Z202472018).

\bibliographystyle{unsrtnat}

\bibliography{references}

\clearpage
\appendix

\begin{center}
	{\LARGE \bf ContinuLoc: Continuous Pose Inference over Neural Fields for UAV Geo-Localization \par}
	\vspace{1em}
	{\Large Supplementary Material \par}
\end{center}
\vspace{2em}

\section{Detailed Dataset Descriptions}
\label{sec:dataset_details}
\subsection{Dataset Collection}
We introduce a novel UAV-satellite image pairing and geo-localization dataset with more accurate geospatial metadata than existing benchmarks. The dataset comprises 12 large-scale scenes, of which 10 are sourced from the UAV-VisLoc~\cite{Xu2024UAVVisLocAL} and 2 are newly collected from publicly available UAV imagery provided by Wingtra, covering a diverse range of topographies including urban areas, croplands, deserts, hilly terrain, and forests.

\begin{figure}[htbp]
	\centering
	\includegraphics[width=\textwidth]{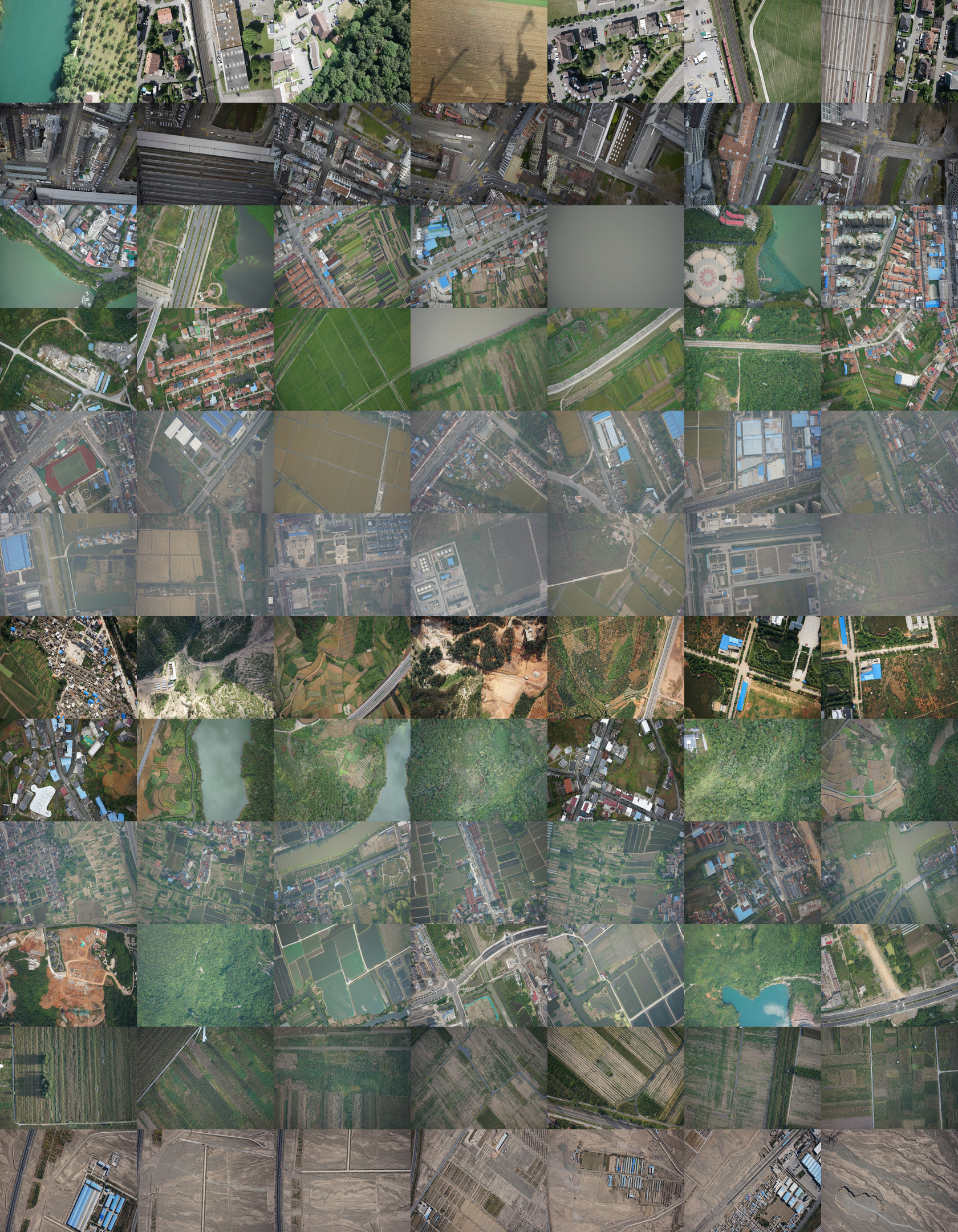}
	\caption{Visual overview of our ContinuScenes cross-view localization dataset. Sample UAV-view images across 12 evaluation scenes. The dataset covers diverse terrains (e.g., urban, farmland, deserts), demonstrating substantial variations in flight altitude and azimuth critical for robust cross-view registration.}
	\label{fig:dataset_overview}
\end{figure}

The absolute altitudes in our dataset encompass an extensive range from 30 m to approximately 2,570 m, while UAV headings exhibit an omnidirectional distribution. Together, these factors provide significant viewpoint and scale diversity. Each UAV-view image is paired with a corresponding reference image crop and annotated with refined geospatial metadata, comprising GPS coordinates, ground coverage, and the heading angle. As illustrated in Figure~\ref{fig:dataset_overview}, sampled UAV-view images exhibit substantial visual variation across the 12 scenes, primarily attributable to differences in flight altitude and heading angle.

\begin{table*}[htbp]
	\centering
	\caption{Summary of the ContinuScenes dataset.}
	\label{tab:dataset_summary}
	\setlength{\tabcolsep}{3.0pt}
	\renewcommand{\arraystretch}{1.12}
	\scriptsize
	\resizebox{\textwidth}{!}{
		\begin{tabular}{lccccccccccccc}
			\toprule
			& Scene-1 & Scene-2 & Scene-3 & Scene-4 & Scene-5 & Scene-6 & Scene-7
			& Scene-8 & Scene-9 & Scene-10 & Scene-11 & Scene-12 \\
			\midrule
			Scene / City
			& Zurich & Zuchwil & Jiujiang01 & Jiujiang02 & Taizhou03 & Taizhou04 & Kunming & Ankang & Huzhou01 & Huzhou02 & Beijing & Zhangye \\
			
			UAV images
			& 2413 & 3951 & 817 & 1071 & 768 & 738 & 473 & 344 & 30 & 1033 & 144 & 590 \\

			Map area (km$^2$)
			& 12 & 12 & 20 & 30 & 64 & 52 & 4 & 6 & 54 & 113 & 2 & 34 \\
			\bottomrule
	\end{tabular}	}
\end{table*}

\subsection{Dataset Construction}
To mitigate cross-view registration errors, inaccurate geospatial metadata, and missing or unreliable heading annotations present in existing benchmarks, we develop an automated pipeline for UAV-to-reference image pairing and geospatial attribute refinement. This pipeline consists of two primary stages: deep-feature-based correspondence estimation, and an affine geometry-based metadata recalibration that builds upon localized patch acquisition and coarse alignment. These steps jointly enable the recovery of accurate GPS coordinates, heading angles, and scale information.

\paragraph{Refined Georeferencing via Affine Matching}
Since neither the UAV-VisLoc dataset nor the Wingtra dataset provides pre-paired reference counterparts, we extract GPS-centered patches from a high-resolution base map and roughly align them to the UAV images by incorporating heading data and phase-domain information, thereby narrowing the search space for refined affine matching. For each UAV image, we estimate a scaling factor $s$ such that the reference patch coverage slightly exceeds the UAV field of view, providing a sufficient search margin for subsequent matching. The patch is cropped from the reference base map, centered on the GPS coordinates recorded in the original flight logs. If heading information is available, we optionally pre-rotate the UAV image to approximately align its heading with the North-aligned reference patch, thereby reducing the rotational search space and facilitating more reliable feature matching.

\paragraph{Affine Matching and Geospatial Metadata Recalibration}

\begin{figure}[htbp]
	\centering
	\includegraphics[width=\linewidth]{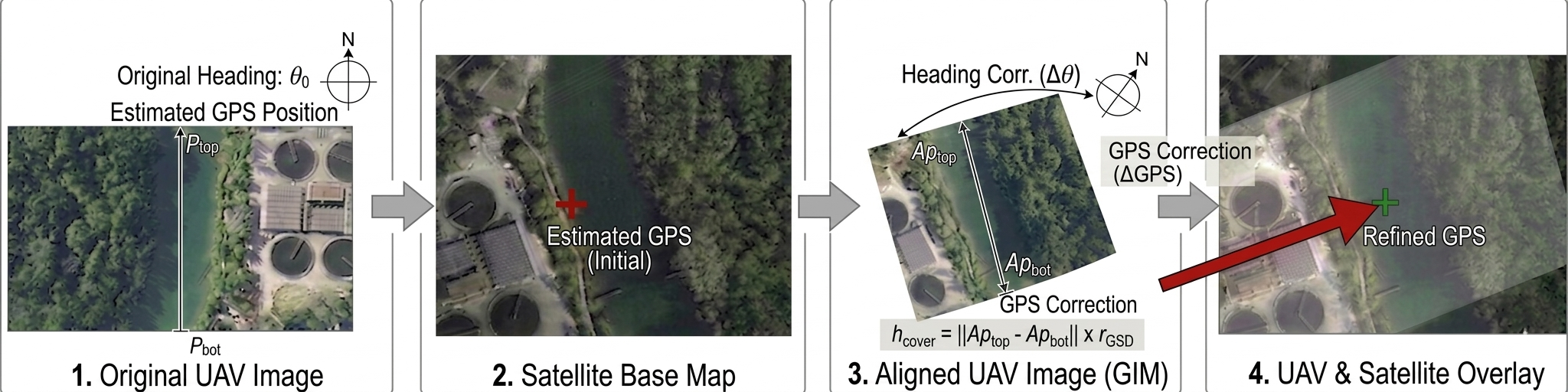}
	\caption{Overview of the UAV and satellite image alignment process. The framework estimates an affine transformation to map the original UAV image (1) to the satellite base map (2). The resulting alignment parameters (3) explicitly drive the recalibration of the UAV's heading ($\Delta\theta$), GPS position ($\Delta\text{GPS}$), and physical footprint ($h_{\text{cover}}$), yielding the final geo-registered overlay (4).}
	\label{fig:alignment}
\end{figure}

To establish precise pixel-level correspondences between UAV and reference views, we employ GIM~\cite{Shen2024GIMLG} with DKM weights. An overview of this alignment process and the subsequent metadata recalibration is visually summarized in Figure~\ref{fig:alignment}. For each cross-view pair, we fit a $2 \times 3$ affine transformation matrix $\mathbf{A}$:
\begin{equation}
	\begin{bmatrix} x' \\ y' \end{bmatrix} 
	= \mathbf{A} \begin{bmatrix} x \\ y \\ 1 \end{bmatrix} 
	= \begin{bmatrix} a & b & t_x \\ c & d & t_y \end{bmatrix} 
	\begin{bmatrix} x \\ y \\ 1 \end{bmatrix}
\end{equation}
where $(x, y)$ and $(x', y')$ denote pixel coordinates in the UAV and reference views, respectively. The linear sub-matrix $\mathbf{M} = \begin{bmatrix} a & b \\ c & d \end{bmatrix}$ encodes rotation and scale, while the translation vector $(t_x, t_y)$ reflects the GPS error in pixel space. Leveraging the estimated $\mathbf{A}$, we recalibrate three geospatial attributes. 

\textbf{(i) Position:} the UAV image center $\mathbf{p}_c = [W_u/2,\, H_u/2,\, 1]^\top$ is projected onto the reference frame via $\mathbf{p}'_c = \mathbf{A}\mathbf{p}_c$, and the resulting pixel displacement relative to the reference patch center is converted into geographic increments to obtain the refined location $(\hat{\text{lat}},\, \hat{\text{lon}})$.

\textbf{(ii) Heading:} the rotational component is extracted from $\mathbf{M}$ via SVD, $\mathbf{M} = \mathbf{U}\boldsymbol{\Sigma}\mathbf{V}^\top$, yielding the nearest rotation matrix $\mathbf{R} = \mathbf{U}\mathbf{V}^\top$. The angular offset $\theta = \operatorname{atan2}(R_{10},\, R_{00})$ is then combined with the raw flight heading and normalized to $[-180^\circ, 180^\circ]$.

\textbf{(iii) Ground Coverage and Scale:} The absolute spatial footprint of each UAV image is quantified by its ground coverage $h_{\text{cover}}$. This metric represents the actual physical extent of the UAV image projected onto the reference map space, defined as:
\begin{equation}
	h_{\text{cover}} = \left\| \mathbf{A}\mathbf{p}_{\text{top}} - \mathbf{A}\mathbf{p}_{\text{bot}} \right\| \times r
\end{equation}
where $\mathbf{A}$ is the ground-truth affine matrix, $\mathbf{p}_{\text{top}}$ and $\mathbf{p}_{\text{bot}}$ denote the centers of the top and bottom edges of the UAV image, and $r$ represents the Ground Sampling Distance (GSD) of the base map. Based on this absolute measurement, we explicitly formulate the relative \textbf{scale factor} as the ratio $h_{\text{cover}} / h_{\text{mean}}$. This conversion from physical coverage to a normalized scale factor enables fine-grained, scale-aware performance analysis across different flight altitudes. Finally, to maintain strict benchmark integrity, any corresponding pairs exhibiting extreme geometric anomalies are systematically curated out.

\subsection{Analysis of Metadata Corrections}
\begin{figure}[h]
	\centering
	\includegraphics[width=\linewidth]{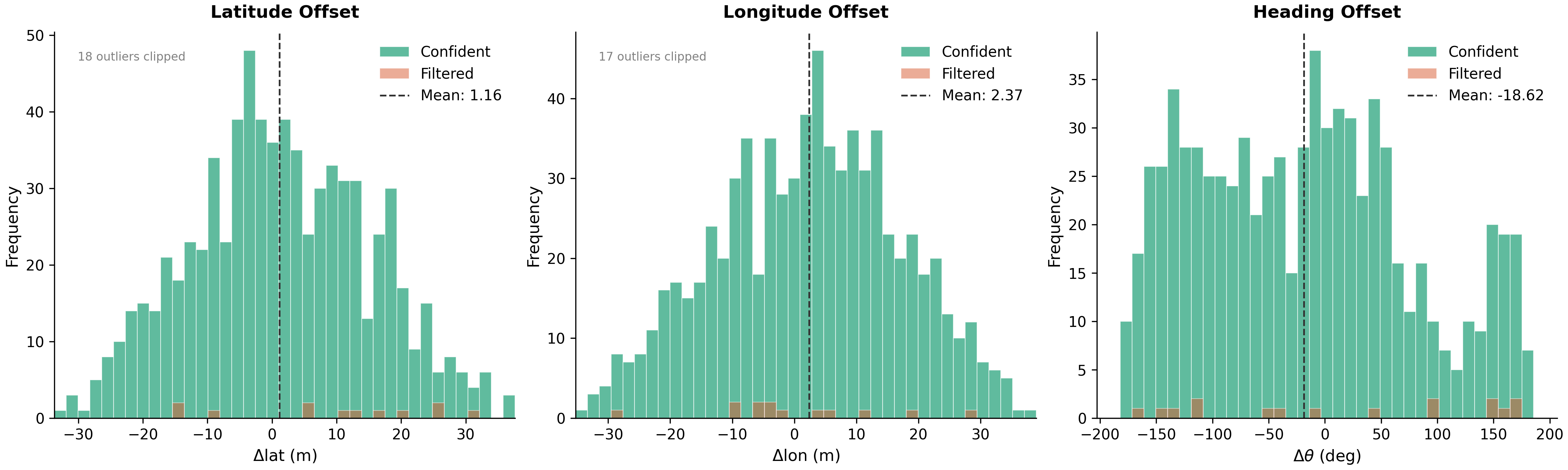}
	\caption{Distribution of metadata corrections ($\Delta\text{lat}$, 
		$\Delta\text{lon}$, $\Delta\theta$) for Taizhou03. 
		Green bars denote confident samples retained after outlier rejection; 
		orange bars denote filtered low-confidence samples.}
	\label{fig:correction_visloc03}
\end{figure}

To quantify the impact of our refinement pipeline, we analyze the distribution of corrections applied to GPS coordinates and heading angles on Taizhou03, as illustrated in Figure~\ref{fig:correction_visloc03}.

The latitude and longitude offsets, $\Delta\text{lat}$ and $\Delta\text{lon}$, reflect the displacement between the original GPS-recorded positions and the affine-corrected locations. While both distributions are approximately centered near zero, the corrections span a considerable range, demonstrating that instance-level positional errors can be substantial and are not adequately captured by the mean alone. The heading offset $\Delta\theta$ exhibits a notably wider distribution, indicating that raw flight headings recorded in the flight 
logs can deviate significantly from the true North-relative orientation, and underscoring the necessity of our affine-based heading correction. In this scene, a small proportion of samples are identified as low-confidence and excluded during the outlier rejection step, confirming the effectiveness of our quality filtering.

Collectively, these results demonstrate that raw geospatial metadata in 
existing UAV datasets can contain non-negligible positional and angular errors, and validate the necessity of our automated refinement pipeline for producing reliable annotations.

\section{Hierarchical Evolutionary Search: Algorithmic Details}
\label{sec:hierarchical_search_details}
In this section, we provide the detailed formulation and algorithmic implementation of our hierarchical evolutionary search procedure, which was outlined in Section~\ref{Hierarchical Pose Inference in Continuous Pose Space}.  
As described in the main text, the search decomposes continuous 4DoF UAV localization into three stages: 
(1) \emph{Coarse Exploration} to obtain a broad coverage of the compatibility landscape, 
(2) \emph{Population Contraction} to iteratively refine local candidate regions, and 
(3) \emph{Local Refinement} using a covariance-adaptive evolutionary optimizer for final pose estimation. 
Here, we provide a detailed description of the symbols, parameters, and update rules used in each stage of the hierarchical search procedure; the corresponding pseudocode is presented in Algorithm~\ref{alg:hds}.

\begin{algorithm}[htbp]
	\caption{Hierarchical Evolutionary Search}
	\label{alg:hds}
	\small
	\DontPrintSemicolon
	\KwIn{compatibility score $S(\mathbf{p})$, pose space $\mathcal{P}$}
	\KwOut{optimal pose estimate $\hat{\mathbf{p}}$}
	
	\BlankLine
	\textbf{Stage I: Coarse Exploration}\\
	Draw a finite support 
	$\mathcal{H}=\{\mathbf{p}_i\}_{i=1}^{N}\sim\mathrm{Unif}(\mathcal{P})$\;
	$\gamma_0 \leftarrow \operatorname{Quantile}_{1-\rho_0}(S(\mathcal{H}))$\;
	$\mathcal{A}_0=\{\mathbf{a}_k\}_{k=1}^{K_0}
	\leftarrow
	\operatorname{Top}_{K_0}
	\{\mathbf{p}\in\mathcal{H}:S(\mathbf{p})\ge\gamma_0\}$\;
	
	\BlankLine
	\textbf{Stage II: Particle-based Population Contraction}\\
	Initialize local populations around $\mathbf{a}_k\in\mathcal{A}_0$:
	$\mathbf{m}_k^{(0)}\leftarrow\mathbf{a}_k$,
	$\delta_k^{(0)}\leftarrow\eta r_{\mathbf{a}_k}$\;
	
	\For{$t=1,\dots,T_{\mathrm{II}}$}{
		\ForEach{active population $k$}{
			Draw a particle set
			$
			\mathcal{X}_k^{(t)}
			=
			\{\mathbf{p}_{k,i}^{(t)}\}_{i=1}^{M}
			\subset
			\mathcal{B}(\mathbf{m}_k^{(t-1)},\delta_k^{(t-1)})
			$
			and evaluate $S(\mathbf{p})$\;
			
			$\gamma_k^{(t)}
			\leftarrow
			\operatorname{Quantile}_{1-\rho}
			\bigl(S(\mathcal{X}_k^{(t)})\bigr)$\;
			
			$\mathcal{E}_k^{(t)}
			\leftarrow
			\{\mathbf{p}\in\mathcal{X}_k^{(t)}:
			S(\mathbf{p})\ge\gamma_k^{(t)}\}$\;
			
			$\mathbf{m}_k^{(t)}
			\leftarrow
			\frac{1}{|\mathcal{E}_k^{(t)}|}
			\sum_{\mathbf{p}\in\mathcal{E}_k^{(t)}}\mathbf{p}$,
			$\delta_k^{(t)}\leftarrow\delta_k^{(0)}\alpha^t$\;
		}
		 Prune populations $k$ whose population-center score $S(\mathbf{m}_k^{(t)})$ falls below the $(1-\rho_{\text{pop}})$ quantile across all active populations\;
	}
	$\mathcal{A}_1
	=
	\{\mathbf{m}_k^{(T_{\mathrm{II}})}\}_{k=1}^{K_1}$,
	$K_1\ll K_0$\;
	
	\BlankLine
	\textbf{Stage III: Parametric Local Refinement}\\
	Select $K_2$ modes from $\mathcal{A}_1$\;
	\ForEach{selected mode $\mathbf{m}_k^{(T_{\mathrm{II}})}$}{
		Initialize local optimizer:
		$
		\tilde{\mathbf{m}}_k^{(0)}
		\leftarrow
		\mathbf{m}_k^{(T_{\mathrm{II}})}
		$,
		$
		\tilde{\delta}_k^{(0)}
		\leftarrow
		\delta_k^{(T_{\mathrm{II}})}
		$,
		$
		\tilde{\mathbf{C}}_k^{(0)}
		\leftarrow
		\mathbf{I}
		$\;
		
		Refine a candidate $\tilde{\mathbf{p}}_k$ by locally maximizing $S(\mathbf{p})$ with covariance-adaptive updates\;
	}
	$\mathcal{A}_2=\{\tilde{\mathbf{p}}_k\}_{k=1}^{K_2}$,
	$K_2\le K_1$\;
	\BlankLine
	$\hat{\mathbf{p}}
	\leftarrow
	\arg\max_{\mathbf{p}\in\mathcal{A}_2}S(\mathbf{p})$\;
	\Return $\hat{\mathbf{p}}$\;
\end{algorithm}

\paragraph{Stage I: Coarse Prior Sampling}
We first draw a finite support $\mathcal{H} = \{\mathbf{p}_i\}_{i=1}^{N}$ from a uniform prior over the 4DoF pose space and evaluate each hypothesis using the compatibility score $S(\mathbf{p})$. 
To retain only high-compatibility hypotheses, we compute a quantile threshold
\begin{equation}
	\gamma_0 = \operatorname{Quantile}_{1-\rho_0}(S(\mathcal{H})),
\end{equation}
which corresponds to the score such that the top fraction $\rho_0$ of hypotheses in $\mathcal{H}$ have compatibility scores above $\gamma_0$. 
From the hypotheses exceeding this threshold, we select $K_0$ coarse candidate anchors:
\begin{equation}
	\mathcal{A}_0=\{\mathbf{a}_k\}_{k=1}^{K_0}.
\end{equation}
Each anchor $\mathbf{a}_k$ is treated as the center of a coarse candidate region and is used to initialize a local population.

\paragraph{Stage II: Population Contraction}
Around each anchor $\mathbf{a}_k \in \mathcal{A}_0$, we initialize a local particle population centered at $\mathbf{m}_k^{(0)} = \mathbf{a}_k$ with an initial sampling radius
$\delta_k^{(0)} = \eta \cdot r_{\mathbf{a}_k}$, where $r_{\mathbf{a}_k}$ is the radius of the local region inherited from Stage~I, defined as half the distance to the nearest neighboring sample in $\mathcal{H}$ and scaled by the factor $\eta$. 
This defines a local neighborhood $\mathcal{B}(\mathbf{m}_k^{(0)},\delta_k^{(0)})$ as an axis-aligned box of half-width $\delta_k^{(0)}$ centered at $\mathbf{m}_k^{(0)}$, from which the initial particles $\mathcal{X}_k^{(0)} = \{\mathbf{p}_{k,i}^{(0)}\}_{i=1}^{M}$ are drawn.
Then, all populations are iteratively refined over $T_{\mathrm{II}}$ rounds. Each refinement round operates at two levels.

Within each population, particles are sampled from the current neighborhood $\mathcal{B}(\mathbf{m}_k^{(t-1)},\delta_k^{(t-1)})$ and evaluated using the compatibility score $S(\mathbf{p})$. 
A threshold $\gamma_k^{(t)} = \operatorname{Quantile}_{1-\rho}\bigl(S(\mathcal{X}_k^{(t)})\bigr)$ is used to select the top-scoring fraction of particles, forming the local elites $\mathcal{E}_k^{(t)}$. 
The population center and sampling radius are updated at each round $t$ as:
\begin{equation}
	\mathbf{m}_k^{(t)} = \frac{1}{|\mathcal{E}_k^{(t)}|} \sum_{\mathbf{p}\in\mathcal{E}_k^{(t)}} \mathbf{p}, \qquad
	\delta_k^{(t)} = \delta_k^{(0)} \cdot \alpha^t,
\end{equation}
where $\mathcal{E}_k^{(t)}$ denotes the set of local elite particles whose scores $S(\mathbf{p})$ exceed the quantile threshold $\gamma_k^{(t)}$, i.e., the top fraction $\rho$ of particles in $\mathcal{X}_k^{(t)}$. Here, $\delta_k^{(0)} = \eta \, r_{\mathbf{a}_k}$ is the initial neighborhood radius inherited from Stage~I, and $\alpha \in (0,1)$ controls the contraction rate of the sampling neighborhood.

Across populations, each population's survival is determined by the score at the mean of its elite particles, and populations whose center score falls below the $(1-\rho_{\text{pop}})$ quantile are pruned.
By combining per-population contraction with cross-population pruning, Stage~II reduces the initial candidate regions to a compact set of retained modes:
\begin{equation}
	\mathcal{A}_1 = \{\mathbf{m}_k^{(T_{\mathrm{II}})}\}_{k=1}^{K_1}, \qquad K_1 \ll K_0.
\end{equation}

\paragraph{Stage III: Local Parametric Refinement}
After Stage~II, the search has been narrowed from the coarse candidate anchors $\mathcal{A}_0$ to a smaller set of retained local modes $\mathcal{A}_1$, each representing a plausible local basin where a single-distribution assumption is reasonable. 
Since $K_1 \ll K_0$, applying local parametric distribution updates to the retained modes is computationally affordable.

We then select at most $K_2 \le K_1$ modes from $\mathcal{A}_1$ and apply a covariance-adaptive evolutionary optimizer to each mode. 
For the $k$-th selected mode, the optimizer is initialized with the population center and radius from Stage~II:
\begin{equation}
	\tilde{\mathbf{m}}_k^{(0)} \leftarrow \mathbf{m}_k^{(T_{\mathrm{II}})}, \quad
	\tilde{\delta}_k^{(0)} \leftarrow \delta_k^{(T_{\mathrm{II}})}, \quad
	\tilde{\mathbf{C}}_k^{(0)} \leftarrow \mathbf{I},
\end{equation}
where $\tilde{\mathbf{C}}_k^{(0)}$ is the initial covariance, set to identity, and subsequently adapted during optimization.

Each optimizer locally maximizes the compatibility score $S(\mathbf{p})$, producing a refined candidate 
$\tilde{\mathbf{p}}_k \in \mathcal{A}_2 = \{\tilde{\mathbf{p}}_k\}_{k=1}^{K_2}$, after which the final estimated pose is
\begin{equation}
	\hat{\mathbf{p}} = \arg\max_{\mathbf{p}\in\mathcal{A}_2} S(\mathbf{p}).
\end{equation}

\section{Experimental Setup and Implementation Details}
\subsection{Implementation Details} \label{sec:implementation}
\paragraph{Visual representation.} 
We use DINOv2 \cite{Oquab2023DINOv2LR} as the visual backbone and follow the parallel-adapter strategy of SeLaVPR~\cite{Lu2024TowardsSA} to fine-tune the last four transformer blocks.
For global aggregation, we use SALAD \cite{Izquierdo2023OptimalTA} with $K=16$ cluster centers of dimension $D=64$, yielding a $K \times D$-dimensional descriptor of size 1024 for each UAV observation and each pose-conditioned reference crop.

\paragraph{Semantic neural field.}
For each scene, we instantiate a multi-resolution hash-grid memory and a lightweight MLP-based decoder that maps retrieved memory to a pose-conditioned semantic descriptor.
The 4DoF pose $\mathbf{p}$ is encoded and injected into a residual-style MLP via feature-wise affine modulation, following the FiLM design \cite{Perez2017FiLMVR}. The residual-style MLP structure is commonly used in neural rendering frameworks \cite{Niemeyer2019DifferentiableVR,Hong2022NeuralMF}.

\paragraph{Hierarchical Evolutionary Search.}
Although Stage~I scores a sparse 4DoF grid, we do not carry the discrete scale dimension into the later evolutionary search. 
For each spatial-heading anchor $\mathbf{a}=(x,y,\theta)$, we collect its compatibility scores over the sampled scale bins $\{s_j\}$ and compute a soft scale estimate
\begin{equation}
\pi_j(\mathbf{a}) =
\frac{\exp(\beta S_j(\mathbf{a}))}
{\sum_l \exp(\beta S_l(\mathbf{a}))},
\qquad
\hat{s}(\mathbf{a}) =
\exp\left(\sum_j \pi_j(\mathbf{a})\log s_j\right),
\end{equation}
where $S_j(\mathbf{a})=\mathcal{C}(\mathbf{z}_q,\phi_{\Theta_M}((x,y,\theta,s_j)))$.
The resulting non-grid scale estimate can be directly carried into subsequent inference because the map field is continuously queryable and admits hypotheses at arbitrary poses beyond the initial discrete grid, a property that standard retrieval libraries do not possess. 
Empirically, this weighted estimate provides more stable scale initialization than a na\"ive $\arg\max$ over the discrete bins. 
We therefore keep $\hat{s}$ fixed in Stage~II and Stage~III, and allocate the evolutionary search budget entirely to the remaining $(x,y,\theta)$ dimensions.

\subsection{Experimental Setup}
\label{sec:Experimental Setup}
\paragraph{Dataset Partitioning}
For training and evaluation, we provide two types of data splits for each scene: 
\textbf{cross-area} splits, which use the first portion of the UAV trajectory for training and the latter portion for testing, and 
\textbf{same-area} splits, which sample queries along the trajectory while maintaining the same-area distribution. 
The corresponding configurations and train/test ratios are summarized in Table~\ref{tab:data_splits}.

\begin{table}[t]
	\centering
	\caption{Data splits for training and evaluation across the four main scenes. Cross-area splits evaluate generalization to unseen trajectory regions, while same-area splits evaluate interpolation along the same trajectory.}
	\label{tab:data_splits}
	\scriptsize
	\setlength{\tabcolsep}{4pt}
	\renewcommand{\arraystretch}{1.1}
	\begin{tabular}{l c c}
		\toprule
		Scene & \shortstack{Cross-Area\\(train/test)} & \shortstack{Same-Area\\(train/test)} \\
		\midrule
		Taizhou03  & 0.8 / 0.2 & 0.8 / 0.2 \\
		Taizhou04  & 0.8 / 0.2 & 0.8 / 0.2 \\
		Zurich     & 0.9 / 0.1 & 0.9 / 0.1 \\
		Zuchwil    & 0.9 / 0.1 & 0.9 / 0.1 \\
		\bottomrule
	\end{tabular}
\end{table}

\paragraph{Discrete pose sampling configuration.}
For the initial finite support $\mathcal{H}$ used in Stage~I, we construct a deterministic grid as a finite-support approximation to a uniform prior over the 4DoF pose space.
For the spatial dimensions, the reference map is tiled into candidate crops according to the image crop size, with zero overlap between adjacent crops; the crop centers define the sampled $(x,y)$ locations. 
For the heading dimension, we sample the full range $[0^\circ,360^\circ)$ at a $10^\circ$ interval, resulting in 36 heading bins. 
For the scale dimension, we first estimate the footprint-size range of UAV images in each scene, and then uniformly sample four scale values within this range. 
The Cartesian product of the spatial grid, heading bins, and scale bins forms the initial hypothesis set $\mathcal{H}$ evaluated in Stage~I:
\begin{equation}
	\mathcal{H}
	=
	\mathcal{G}_{xy}
	\times
	\Theta
	\times
	\mathcal{S},
	\qquad
	\Theta=\{0^\circ,10^\circ,\ldots,350^\circ\},
	\quad
	|\mathcal{S}|=4.
\end{equation}
For the discrete retrieval-based 4DoF baselines in Table~\ref{tab:main_res_interval}, we use the same sampling configuration to construct their reference libraries. 
For the 2DoF retrieval-based baselines, only the spatial grid is used: the heading is fixed to $0^\circ$, and the scale is set to the scene-specific mean UAV footprint size.

For the sampling-density experiment in Section~\ref{sec:Comparison with Dense Retrieval}, we progressively increase the spatial sampling density by raising the overlap ratio between adjacent map crops, starting from zero overlap. 
The heading and scale dimensions are kept unchanged, with 36 heading bins and 4 scale bins, respectively.

\paragraph{Ablation configuration.}
For the stage-wise inference ablation experiment in Table~\ref{tab:stagewise_progression}, we retain the top $5\%$ of hypotheses from Stage~I as the initialization set for Stage~II, and further select the top-5 retained modes from Stage~II for Stage~III refinement. 
This configuration is kept consistent with the inference setting used in the main results in Table~\ref{tab:main_res_interval}.
As shown in Figure~\ref{fig:dense_retrieval_tradeoff}, Stage~III accounts for most of the inference time, since covariance-adaptive refinement must maintain and update parametric statistics, including the mean, step size, and covariance matrix, for each selected mode. 
This motivates our hierarchical design: a parameter-light population contraction stage first reduces the number of plausible modes, so that the expensive covariance-adaptive optimizer is applied only to a few high-compatibility candidates.

\paragraph{Training Configuration }
Table~\ref{tab:training_compact} shows the main training configurations for Visual Representation Learning and Map Representation Learning, including optimizer, learning rate, weight decay, and number of epochs.
Visual Representation Learning trains the visual aggregation head and adapters with metric learning, while Map Representation Learning freezes the visual encoder and trains the map field to regress L2-normalized visual descriptors.
All experiments were performed on Ubuntu 20.04.6 LTS using a single NVIDIA RTX 4090 GPU with 24\,GB of memory, with CPU computations limited to a single thread by default.
\begin{table}[htbp]
	\centering
	\scriptsize
	\setlength{\tabcolsep}{6pt}
	\renewcommand{\arraystretch}{1.1}
	\caption{Training configurations for Visual Representation Learning and Map Representation Learning.}
	\label{tab:training_compact}
	\begin{tabular}{l l l l}
		\toprule
		Module & Trainable & Optimizer & LR / WD / Epochs \\
		\midrule
		Visual Representation Learning & Aggregator / Adapters & Adam & 1e-4 / 1e-6 / 100 \\
		Map Representation Learning & Semantic Neural Field & Adam & 1e-2 / 1e-6 / 1000 \\
		\bottomrule
	\end{tabular}
\end{table}

\section{Additional Experiments}
\begin{table*}[htbp]
	\centering
	\caption{Comparison of representative localization paradigms on four scenes in the cross-area setting.}
	\setlength{\tabcolsep}{3.2pt}
	\renewcommand{\arraystretch}{1.12}
	\scriptsize
	\label{tab:main_res_interval_cross_area}
	
	\begin{tabular}{lcccccccccccc}
		\toprule
		& \multicolumn{6}{c}{Taizhou03}
		& \multicolumn{6}{c}{Taizhou04} \\
		\cmidrule(lr){2-7} \cmidrule(lr){8-13}
		& \multicolumn{3}{c}{R@1 (100m, 10$^\circ$, 1.2x) $\uparrow$}
		& \multicolumn{3}{c}{Median@Success $\downarrow$}
		& \multicolumn{3}{c}{R@1 (100m, 10$^\circ$, 1.2x) $\uparrow$}
		& \multicolumn{3}{c}{Median@Success $\downarrow$} \\
		\cmidrule(lr){2-4} \cmidrule(lr){5-7}
		\cmidrule(lr){8-10} \cmidrule(lr){11-13}
		Method
		& D & D\&H & D\&H\&S & Dist & Head & Scale
		& D & D\&H & D\&H\&S & Dist & Head & Scale \\
		\midrule
		DenseUAV~\cite{Dai2022VisionBasedUS}
		& 30.20 & -- & -- & 69.56 & -- & --
		& 16.08  & -- & -- & 73.61 & -- & -- \\
		QDFL~\cite{Hu2025QueryDrivenFL} & 22.67 & -- & -- & 60.76 & -- & --
		& 16.78 & -- & -- & 74.49 & -- & -- \\ 
		\midrule
		NetVLAD~\cite{Arandjelovi2015NetVLADCA}
		& 26.00 & 15.33 & 14.67 & 66.05 & 5.81 & 1.062
		& 14.69 & 11.89 & 11.89 & 74.46 & 3.83 & 1.039 \\
		SALAD~\cite{Izquierdo2023OptimalTA}
		& 22.67 & 17.33 & 17.33 & 65.21 & 4.24 & 1.106
		& 12.50 & 8.33  & 8.33  & 74.59 & \textbf{2.19} & 1.055 \\
		SeLaVPR~\cite{Lu2024TowardsSA}
		& 26.67 & 18.67 & 18.67 & 66.05 & \textbf{3.75} & 1.070
		& 15.38 & 13.29 & 13.29 & 74.46 & 2.27 & 1.063 \\
		\midrule
		PoseNet~\cite{Kendall2015PoseNetAC}
		& 0.00 & 0.00 & 0.00 & -- & -- & --
		& 0.00 & 0.00 & 0.00 & -- & -- & -- \\
		\midrule
		Ours
	& \textbf{87.33} & \textbf{74.67} & \textbf{74.67} & \textbf{36.30} & 4.48 & \textbf{1.051}
    & \textbf{58.04} & \textbf{53.15} & \textbf{53.15} & \textbf{41.65} & 3.94 & \textbf{1.038} \\
		
		\specialrule{0.08em}{0.4em}{0.4em}
		
		& \multicolumn{6}{c}{Zurich}
		& \multicolumn{6}{c}{Zuchwil} \\
		\cmidrule(lr){2-7} \cmidrule(lr){8-13}
		& \multicolumn{3}{c}{R@1 (50m, 10$^\circ$, 1.2x) $\uparrow$}
		& \multicolumn{3}{c}{Median@Success $\downarrow$}
		& \multicolumn{3}{c}{R@1 (50m, 10$^\circ$, 1.2x) $\uparrow$}
		& \multicolumn{3}{c}{Median@Success $\downarrow$} \\
		\cmidrule(lr){2-4} \cmidrule(lr){5-7}
		\cmidrule(lr){8-10} \cmidrule(lr){11-13}
		Method
		& D & D\&H & D\&H\&S & Dist & Head & Scale
		& D & D\&H & D\&H\&S & Dist & Head & Scale \\
		\midrule
		DenseUAV~\cite{Dai2022VisionBasedUS}
		& 19.83 & -- & -- & 24.21 & -- & --
		& 18.24 & -- & -- & 33.97 & -- & -- \\
		QDFL~\cite{Hu2025QueryDrivenFL}
		& 27.62 & -- & -- & 23.56 & -- & --
		& 24.48 & -- & --  & 27.07 & -- & -- \\
		\midrule
		NetVLAD~\cite{Arandjelovi2015NetVLADCA}
		& 32.22 & 17.99 & 8.79  & 29.49 & 4.46 & 1.099
		& 31.72 & 17.93 & 13.45 & 27.86 & 4.07 & 1.042 \\
		SALAD~\cite{Izquierdo2023OptimalTA}
		& 30.54 & 14.23 & 9.62  & 29.31 & 3.71 & 1.114
		& 30.00 & 16.21 & 13.45 & 26.20 & 5.13 & 1.040 \\
		SeLaVPR~\cite{Lu2024TowardsSA}
		& 28.87 & 16.74 & 10.04 & 29.49 & \textbf{3.07} & 1.105
		& 23.45 & 13.10 & 13.10 & 28.78 & 4.37 & \textbf{1.025} \\
		\midrule
		PoseNet~\cite{Kendall2015PoseNetAC}
		& 0.00 & 0.00 & 0.00 & -- & -- & --
		& 2.07 & 0.34 & 0.34 & 33.80 & 8.76 & 1.187 \\
		\midrule
		Ours
	& \textbf{84.94} & \textbf{65.27} & \textbf{62.76} & \textbf{9.80} & 3.10 & \textbf{1.063}
	& \textbf{45.86} & \textbf{33.45} & \textbf{33.45} & \textbf{13.57} & \textbf{3.75} & 1.159 \\
		\bottomrule
	\end{tabular}
\end{table*}

\begin{table*}[tbp]
	\centering
	\caption{Effect of GIM$_{dkm}$-based local refinement on progressive recall under the cross-area setting.}
	\label{tab:gim_refinement_progressive_recall_segment}
	\setlength{\tabcolsep}{2.6pt}
	\renewcommand{\arraystretch}{0.98}
	\scriptsize
	\begin{tabular}{lcccccccccccc}
		\toprule
		& \multicolumn{3}{c}{Taizhou03}
		& \multicolumn{3}{c}{Taizhou04}
		& \multicolumn{3}{c}{Zurich}
		& \multicolumn{3}{c}{Zuchwil} \\
		\cmidrule(lr){2-4}
		\cmidrule(lr){5-7}
		\cmidrule(lr){8-10}
		\cmidrule(lr){11-13}
		Method
		& D & D\&H & D\&H\&S
		& D & D\&H & D\&H\&S
		& D & D\&H & D\&H\&S
		& D & D\&H & D\&H\&S \\
		\midrule
		
		QDFL~\cite{Hu2025QueryDrivenFL}
		& 72.67 & 72.67 & 72.67
		& 63.64 & 63.64 & 62.94
		& 54.39 & 53.97 & 48.12
		& 48.62 & 47.93 & 45.52 \\
		
		SeLaVPR~\cite{Lu2024TowardsSA}
		& 75.33 & 75.33 & 75.33
		& 51.05 & 51.05 & 50.35
		& 56.90 & 55.23 & 48.54
		& 45.86 & 45.86 & 44.48 \\
		
		Ours
		& \textbf{98.00} & \textbf{98.00} & \textbf{98.00}
		& \textbf{81.12} & \textbf{81.12} & \textbf{81.12}
		& \textbf{89.12} & \textbf{89.12} & \textbf{83.26}
		& \textbf{60.69} & \textbf{59.31} & \textbf{57.59} \\
		
		\bottomrule
	\end{tabular}
\end{table*}

\begin{figure}[htbp]
	\centering
	\includegraphics[width=0.75\linewidth]{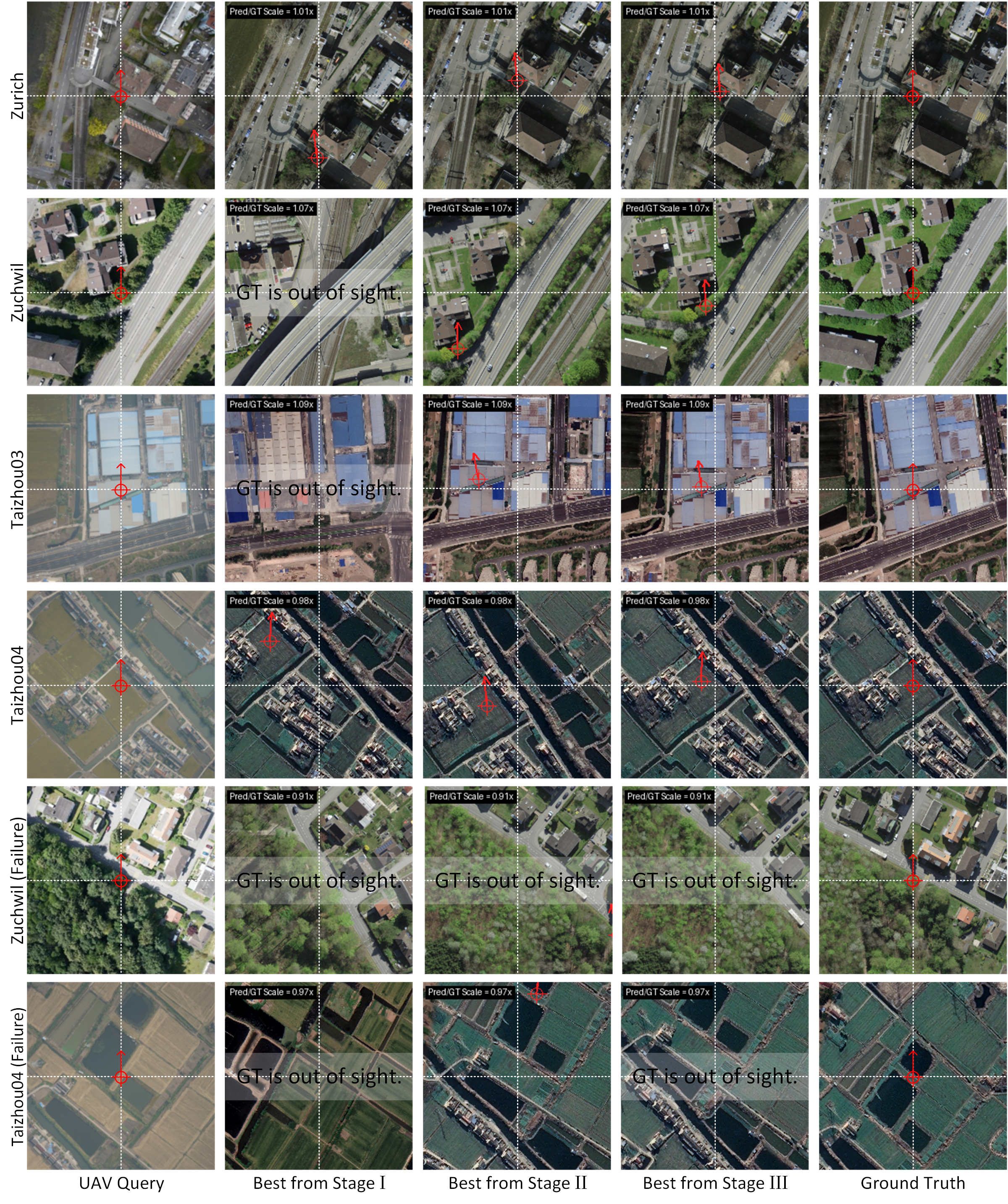}
	\caption{ Qualitative visualization of stage-wise pose refinement in the cross-area setting.
	The red arrows mark the relative positions and orientations of the ground-truth query poses within the coordinate frames of the hypothesized reference views.}
	\label{fig:stagewise_progression_cross}
\end{figure}

\paragraph{Main Results under the Cross-area Setting}
As shown in Table~\ref{tab:main_res_interval_cross_area}, we present the performance of our method under the cross-area setting. Compared to its performance in the same-area setting (Table~\ref{tab:main_res_interval}), our continuous 4DoF inference largely maintained recall and precision on Taizhou03 and Zurich, indicating that the method possesses strong cross-region generalization. Although some performance drop occurs in Taizhou04 and Zuchwil, this trend is not unique to our approach: most baseline methods also exhibit decreased recall and precision in these scenes, suggesting that the decline is primarily driven by scene-specific factors. Nevertheless, our method consistently achieves the highest R@1 under all three criteria and the lowest median position error among successful localizations across all four scenes. 
Furthermore, after applying GIM-based~\cite{Shen2024GIMLG} local refinement, our method achieves the highest R@1 under all three criteria across all four scenes, as shown in Table~\ref{tab:gim_refinement_progressive_recall_segment}. 
Qualitative results in the cross-area setting are shown in Figure~\ref{fig:stagewise_progression_cross}.

\paragraph{Detailed Analysis of Dense Retrieval Libraries.} Table~\ref{tab:appendix_dense_retrieval_recall1} further examines the impact of reference library density, complementing Figure~\ref{fig:dense_retrieval_tradeoff}. Under the full $D\&H\&S$ criterion, increasing the sampling density improves dense-retrieval R@1 from 7.98\% to 36.13\% on Zurich and from 21.48\% to 73.15\% on Taizhou03. In comparison, our continuous inference achieves R@1 of 70.59\% and 81.21\%, respectively, with 98k initial hypotheses per scene. These results demonstrate higher recall than the densest retrieval configurations while using substantially fewer initial hypotheses.

\begin{table*}[htbp]
	\centering
	\caption{
		Recall@1 of dense retrieval under increasing gallery density on Zurich and Taizhou03.
	}
	\label{tab:appendix_dense_retrieval_recall1}
	\scriptsize
	\setlength{\tabcolsep}{3.2pt}
	\renewcommand{\arraystretch}{1.06}
	\begin{tabular}{llcccccccc}
		\toprule
		\multirow{2}{*}{Paradigm}
		& \multirow{2}{*}{Overlap}
		& \multicolumn{4}{c}{Zurich R@1 (50m, 10$^\circ$, 1.2x)}
		& \multicolumn{4}{c}{Taizhou03 R@1 (100m, 10$^\circ$, 1.2x)} \\
		\cmidrule(lr){3-6} \cmidrule(lr){7-10}
		&
		& Gallery $\downarrow$
		& $D$ $\uparrow$
		& $D\&H$ $\uparrow$
		& $D\&H\&S$ $\uparrow$
		& Gallery $\downarrow$
		& $D$ $\uparrow$
		& $D\&H$ $\uparrow$
		& $D\&H\&S$ $\uparrow$ \\
		\midrule
		
		\multirow{8}{*}{\makecell[l]{Dense Ret.}}
		& 0.000 
		& 98k   & 29.41 & 20.17 & 7.98
		& 98k   & 28.19 & 22.82 & 21.48 \\
		& 0.125 
		& 135k  & 31.93 & 23.11 & 10.50
		& 120k  & 32.89 & 24.83 & 22.15 \\
		& 0.250 
		& 187k  & 39.50 & 31.09 & 16.39
		& 160k  & 40.94 & 32.89 & 30.87 \\
		& 0.375 
		& 259k  & 55.04 & 42.86 & 19.33
		& 240k  & 60.40 & 46.98 & 43.62 \\
		& 0.500 
		& 408k  & 67.65 & 54.62 & 23.95
		& 378k  & 75.17 & 61.74 & 54.36 \\
		& 0.675 
		& 954k  & 79.83 & 68.07 & 30.25
		& 900k  & 88.59 & 75.84 & 70.47 \\
		& 0.750 
		& 1.65M & 85.71 & 73.95 & 31.09
		& 1.51M & 90.60 & 77.85 & 71.14 \\
		& 0.875 
		& 6.57M & 89.92 & 79.41 & 36.13
		& 5.98M & 90.60 & 79.87 & 73.15 \\
		
		\specialrule{0.08em}{0.4em}{0.4em}
		
		Ours
		& 0.000
		& \textbf{98k} & \textbf{89.92} & \textbf{81.93} & \textbf{70.59}
		& \textbf{98k} & \textbf{91.28} & \textbf{81.21} & \textbf{81.21} \\
		
		\bottomrule
	\end{tabular}
\end{table*}

\section{Limitations} 
\label{sec:limitation}
Despite achieving strong performance compared to existing paradigms, our method has several limitations: 
(1) Constructing a continuous map representation for a new map incurs an additional time overhead, ranging from 0.5 to 3 hours, depending primarily on the size of the map.
(2) Fitting a map is, in fact, an underlying scientific challenge; not all feature spaces learned by visual encoders are amenable to such fitting.
(3) Although our dataset encompasses a variety of real-world drone scenarios from different cities, only the "Zurich" and "Zuchwil" scenarios include flights conducted at varying altitudes; in the other scenarios, the range of flight altitudes remains relatively consistent. This is why our inference algorithm does not optimize for scale as the primary variable.

\end{document}